\documentclass{egpubl}
\usepackage{VMV2025}

\ArxivPaper

\usepackage[T1]{fontenc}
\usepackage{dfadobe}  

\usepackage{amsmath}
\usepackage{amsfonts}
\usepackage{xcolor}
\usepackage{microtype}
\usepackage{placeins}

\renewcommand{\thefootnote}{\fnsymbol{footnote}}

\newcommand{\beginsupplement}{%
        \setcounter{table}{0}
        \renewcommand{\thetable}{S.T\arabic{table}}%
        \setcounter{figure}{0}
        \renewcommand{\thefigure}{S.F\arabic{figure}}%
        \renewcommand{\thesection}{S\arabic{section}}%
     }

\usepackage{cite}  
\BibtexOrBiblatex
\electronicVersion
\PrintedOrElectronic

\usepackage{hyperref}
\hypersetup{
    colorlinks=true,
    linkcolor=blue,      
    urlcolor=cyan,
}

\ifpdf \usepackage[pdftex]{graphicx} \pdfcompresslevel=9
\else \usepackage[dvips]{graphicx} \fi

\definecolor{jorange}{rgb}{8,.5,.0}

\usepackage{booktabs}
\usepackage{colortbl}
\usepackage{tikz}
\usepackage{float}

\newcommand\nir{near-infra red}

\def\cg{\cellcolor{green!40}}

\def\cred{\cellcolor{red!40}}

\usepackage{egweblnk} 

\title{Towards Integrating Multi-Spectral Imaging with Gaussian Splatting}
\author[J. Grün\textsuperscript{\textdagger} \& L. Meyer\textsuperscript{\textdagger} et al.]
{\parbox{\textwidth}{\centering Josef Grün\textsuperscript{\textdagger}, Lukas Meyer\textsuperscript{\textdagger} \orcid{0000-0003-3849-7094
}, Maximilian Weiherer \orcid{0009-0001-4615-4684}, Bernhard Egger \orcid{0000-0002-4736-2397}, Marc Stamminger \orcid{0000-0001-8699-3442} and Linus Franke \orcid{0000-0001-8180-0963}
        }
        \\
{\parbox{\textwidth}{\centering Visual Computing Erlangen, Friedrich-Alexander-Universität Erlangen-Nürnberg, Germany
       }
}
}

\begin{document}
\teaser{%
\centering
 \includegraphics[width=0.98\linewidth]{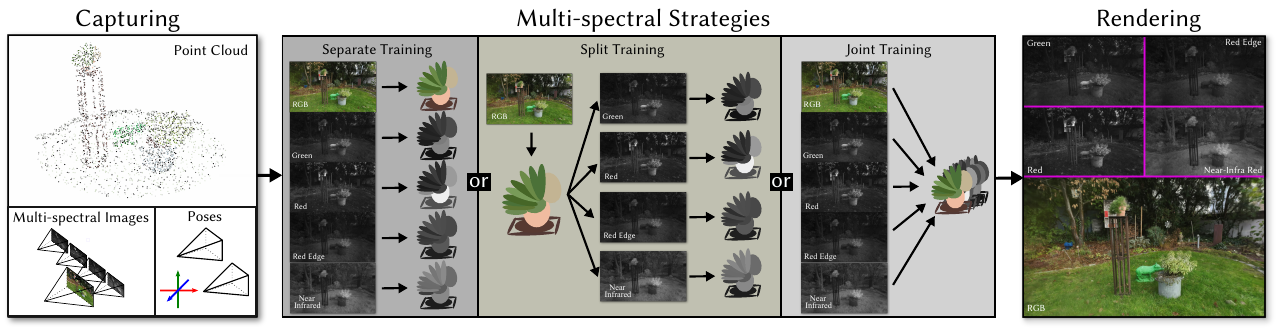}
 \centering
    \caption{We explore the integration of multi-spectral captures containing RGB and further spectral bands (e.g., red edge, near infra-red) into the 3D Gaussian Splatting (3DGS) framework.
    We investigate three main strategies:
   \textsc{Separate} individually optimizes a 3DGS model for each band, whereas \textsc{Split} initially constructs an RGB model, replicates the geometry, and then fits each channel separately through geometry and color optimization; \textsc{Joint} reconstructs a single model jointly, with each primitive holding multiple spectral colors.
}%
    \label{fig:pipeline}
}%

\maketitle

\begin{abstract}
We present a study of how to integrate color (RGB) and multi-spectral imagery (red, green, red-edge, and near-infrared) into the 3D Gaussian Splatting (3DGS) framework -- a state-of-the-art explicit radiance-field-based method for fast and high-fidelity 3D reconstruction from multi-view images~\cite{kerbl3Dgaussians}. 
While 3DGS excels on RGB data, na\"ive per-band optimization of additional spectra yields poor reconstructions due to inconsistently appearing geometry in the spectral domain. 
This problem is prominent, even though the actual geometry is the same, regardless of spectral modality.
To investigate this, we evaluate three strategies: 1) \textup{Separate} per-band reconstruction with no shared structure; 2) \textup{Splitting optimization}, in which we first optimize RGB geometry, copy it, and then fit each new band to the model by optimizing both geometry and band representation; and 3) \textup{Joint}, in which the modalities are jointly optimized, optionally with an initial RGB-only phase.
We showcase through quantitative metrics and qualitative novel-view renderings on multi-spectral datasets the effectiveness of our dedicated optimized \textup{Joint} strategy, increasing overall spectral reconstruction as well as enhancing RGB results through spectral cross-talk.
We therefore suggest integrating multi-spectral data directly into the spherical harmonics color components to compactly model each Gaussian’s multi-spectral reflectance.
Moreover, our analysis reveals several key trade-offs in when and how to introduce spectral bands during optimization, offering practical insights for robust multi-modal 3DGS reconstruction.


The \textbf{project page} and \textbf{code} is located at: \href{https://meyerls.github.io/towards_multi_spec_splat}{\nolinkurl{meyerls.github.io/towards\_multi\_spec\_splat}}


%
\end{abstract}%

\begingroup \renewcommand\thefootnote{}\footnotetext{ \textsuperscript{\textdagger} Equal contribution\\
Email Josef Gr\"un: \texttt{\{firstname\}.m.gruen@fau.de} \\
Email others: \texttt{\{firstname.lastname\}@fau.de}  \\
}        
\endgroup


\section{Introduction}

3D Gaussian Splatting (3DGS)~\cite{kerbl3Dgaussians} has rapidly emerged as the state‑of‑the‑art framework for real‑time radiance field reconstruction, combining fast training, efficient memory usage, and high‐fidelity rendering. By representing a scene as a collection of anisotropic 3D Gaussians with learnable covariances and optimizable colors via spherical harmonics, 3DGS methods avoid the costly sampling of dense volumetric grids or implicit functions while still preserving continuous volumetric radiance properties.\\
\indent In parallel, multi-spectral imaging, capturing bands beyond the visible spectrum such as \nir (NIR) and red‑edge, has become increasingly important in applications ranging from precision agriculture~\cite{Huang2020} to material identification~\cite{namin2012classification}. 
These additional spectral channels expose physical and chemical properties of surfaces that RGB alone cannot reveal, and recent works have begun to extend 3DGS to multi-spectral domains.
Exploiting this modality allows capturing of more information beyond visible light, which can boost reconstruction quality~\cite{meyer2025multi}, an effect termed \textit{spectral cross-talk}.\\
\indent Naively integrating multi-spectral data into existing 3DGS pipelines remains challenging.
Despite the robustness of RGB reconstruction in 3DGS due to dense and feature-rich information in RGB images and a specialized Gaussian densification scheme, using individual spectral channels such as NIR or red‑edge in isolation often produces degraded or unstable reconstructions when processed. 
This discrepancy limits the applicability of 3DGS to true multi-spectral workflows, where balanced quality across all bands is essential.
In this paper, we systematically investigate three methodologies for fusing multi-spectral data into the 3DGS framework:
\begin{enumerate}
    \item \textbf{Independent per‑band optimization (\textsc{Separate}):} Each spectral channel is represented and optimized as a separate 3DGS model with no shared geometry or parameters, serving as a lower‐bound reference for reconstruction quality, prone to artifacts.
    \item \textbf{Sequential split reconstruction (\textsc{Split}):} We first optimize a high‑quality RGB 3DGS model, then \textit{split} the optimization by copying its geometry into multiple band‑specific Gaussian models and refine each band individually. This two‑stage process preserves geometry while adapting appearance to each spectrum.
    \item \textbf{Staged joint optimization (\textsc{Joint}):} We introduce a unified pipeline that optimizes all spectral channels simultaneously. We extend this methodology with training schemes (a) initially optimizing only the RGB channels to establish stable geometry, and (b) selectively freezing or updating structural parameters (positions, covariances) to balance fidelity and cross‑band consistency.
\end{enumerate}

We evaluate these strategies on real multi-spectral data consisting of RGB, narrow-band red, narrow-band green, red‑edge, and NIR captures (7 bands total) and analyze per‑band reconstruction metrics as well as interaction (spectral cross-talking) effects.
Our findings reveal key trade‑offs between geometric coupling and spectral fidelity, and we conclude with practical guidelines for integrating multi-spectral data into 3D Gaussian Splatting systems.
In summary, we contribute:
\begin{itemize}
    \item The formulation and comparison of multi-spectral integration strategies: We introduce and systematically evaluate three optimization paradigms, \textsc{Separate}, \textsc{Split}, and \textsc{Joint}, for incorporating additional spectral bands into 3DGS.
    \item We propose a set of optimizations for the \textsc{Joint} strategy for multi-spectral reconstruction as well as practical insights resulting from this analysis. 
    \item We examine how spectral cross-talk between spectral bands can improve reconstruction quality.
\end{itemize}

\section{Related Work}

\subsection{Novel View Synthesis}
Traditional methods for novel view synthesis rely on image-based rendering~\cite{shum2000review,shum2008image}, where images are re-projected using depth from multi-view stereo~\cite{schoenberger2016mvs} or Structure-from-Motion~\cite{schonberger2016structure}. 
More recently, neural rendering methods~\cite{tewari2020state,tewari2022advances} have enabled end-to-end optimization of geometry and appearance via differentiable rendering. 
Neural Radiance Fields (NeRF)~\cite{mildenhall2020nerf} introduced volumetric scene modeling with implicit MLPs, followed by works improving rendering quality~\cite{zhang2020nerf++,barron2022mipnerf360,hahlbohm2024inpc} and speed~\cite{mueller2022instant,barron2023zip}. 
Neural point-based methods directly render reconstructed point clouds and process renderings with neural networks~\cite{aliev2020neural}, with increasing degrees of point cloud optimizations and fast rendering%
~\cite{trips,vet,you2023learning,inovis}.

3D Gaussian Splatting (3DGS)~\cite{kerbl3Dgaussians} revolutionized the field by replacing volumetric fields with explicit, anisotropic 3D Gaussians, offering high-quality, real-time rendering. Its fast convergence and compact representation have made it a new foundation in neural rendering. Extensions have improved temporal consistency~\cite{luiten2023dynamic,wu20244dgs}, scalability~\cite{kerbl2024hierarchical,lin2024vastgaussian}, performance~\cite{radl2024stopthepop,hahlbohm2025htgs,yu2024mip,niemeyer2024radsplat}, and compression~\cite{3DGSzip2024,Niedermayr_2024_CVPR}. 

\subsection{Multi-modal and Multi-spectral Reconstruction}
Multi-modal methods aim to combine different sensor types, such as reflectance (e.g., RGB, NIR) or emission (e.g., thermal). This paper focuses on the integration of passive multi-spectral reflectance channels into 3DGS. While multi-spectral sensors typically capture a small number of discrete channels, hyper-spectral systems record dense, narrow-band spectra~\cite{Adam2010}.

Recent work explores integrating these modalities into neural reconstruction pipelines. For example, NeRF variants including thermal data~\cite{lin2024thermalnerf,Jiacongthermalnerf,ye2024thermalnerfneuralradiancefields, mert,hassan2024thermonerf} enhance low-light performance by fusing RGB and thermal data.

In the 3DGS domain, Thermal3D-GS~\cite{chen2024thermal3dgs} pioneered thermal-only splatting, while ThermalGaussian~\cite{lu2024thermalgaussianthermal3dgaussian} jointly optimizes RGB and thermal channels using separate spherical harmonic models for each. SpectralGaussians~\cite{sinha2024spectralgaussianssemanticspectral3d} further augments 3DGS with physically-based shading that estimates per-spectrum reflectance and illumination for each Gaussian.

For richer spectral data, Spec-NeRF~\cite{spec-nerf} and X-NeRF~\cite{poggi2022xnerf} integrate multi-spectral bands while estimating camera parameters or sensitivity functions. SpectralNeRF~\cite{spectralnerf} reconstructs per-channel spectra which are later fused. HS-NeRF~\cite{Chen24arxiv_HS-NeRF} extends this to hyper-spectral inputs by conditioning on continuous wavelength embeddings. HyperGS~\cite{HyperGS} adapts this idea to 3DGS by learning a compressed latent spectral representation decoded at render time.


Recent research on multi-spectral radiance fields in Gaussian splatting has introduced a cross-spectral neural color component for each Gaussian~\cite{meyer2025multi}. 
This approach embeds spectral appearance in a high-dimensional feature vector decoded via a small MLP. However, it sacrifices the real-world interpretability of per-Gaussian spherical harmonics in favor of a learned encoding.

These methods highlight the growing need for multi-spectral scene representations and motivate our evaluation of strategies for integrating spectral information into the 3DGS framework.

The methodologies proposed in ThermalGaussian~\cite{lu2024thermalgaussianthermal3dgaussian} (and in \"Ozer et al.~\cite{mert} for NeRF-like methods) are closely related to ours, with a particular emphasis on the integration of RGB and thermal data. In contrast to multi-spectral data, thermal imagery tends to exhibit a smoother appearance due to the relatively homogeneous temperature distribution within local object regions. The multi-spectral data we use provides more fine-grained geometric and structural information, making it particularly well-suited for capturing subtle object details. To explore how these methods generalize to multi-spectral data and to investigate the effects of spectral cross-talk, we also systematically explore the integration space and extend it through additional experimental studies and improvements to the underlying training strategies.


\subsection{Preliminaries: 3D Gaussian Splatting}

3D Gaussian Splatting (3DGS)~\cite{kerbl3Dgaussians} represents a scene using a set of anisotropic 3D Gaussians. By optimizing the position, scale, orientation, color, and opacity of these splats, the appearance of a real-world scene can be efficiently rendered from arbitrary viewpoints.
Starting from a set of RGB images, a colored sparse point cloud and corresponding camera poses are recovered using Structure-from-Motion (SfM)~\cite{schonberger2016structure}. The sparse point cloud is used to initialize a set of 3D Gaussians, where each Gaussian is centered at a position (mean) $\boldsymbol{\mu}$ and its extent is defined by a $3\times 3$ covariance matrix $\boldsymbol{\Sigma}$. The density of a Gaussian distribution at a point $\mathbf{x} \in \mathbb{R}^3$ is given by
$G(\mathbf{x}) = e^{-\frac{1}{2}(\mathbf{x} - \boldsymbol{\mu})^{T} \boldsymbol{\Sigma}^{-1} (\mathbf{x} - \boldsymbol{\mu})} $
and the covariance matrix (which geometrically represents an ellipsoid in 3D space) can be factorized as
$ \boldsymbol{\Sigma} = \mathbf{R} \mathbf{S} \mathbf{S}^\top \mathbf{R}^\top, $
where $\mathbf{S}$ is a diagonal scaling matrix and $\mathbf{R}$ is a rotation matrix.
Thus, when optimizing $\boldsymbol{\Sigma}$, essentially scale $\mathbf{S}$ and rotation $\mathbf{R}$ are refined.
Using a tile-based rasterization pipeline, Kerbl et al.~\cite{kerbl3Dgaussians} project the Gaussian to screen with EWA splatting~\cite{zwicker2001surface} and 
$\alpha$-blend the contributing Gaussians with opacities $\alpha_i$.
Colors in 3DGS are represented with three degrees of spherical harmonics, allowing view-dependent RGB representation.


Importantly, during optimization, the Adaptive Density Control (ADC) module enhances the Gaussian set's density by cloning or splitting primitives based on gradient heuristics and removes outlier Gaussians with opacity thresholding. 
This module is typically active for half the iterations of the optimization, allowing the second half to fine-tune the primitives.

\section{Method}
\label{sec:method}
In this study, we aim to demonstrate how to effectively integrate multi-spectral data into the common 3DGS frameworks~\cite{kerbl3Dgaussians} using different training paradigms.
Additionally, as spectral data reveals more details of the objects and materials in the scene, we investigate how and to what degree different spectral bands can help increase reconstruction quality in the other modalities, an effect termed spectral cross-talk.

We present three different strategies to include spectral data into the 3DGS pipeline: (1) Separate optimization, (2) splitting spectral models and optimization after an initial RGB reconstruction, and (3) jointly optimizing all bands together.
For an overview of these strategies, see Fig.~\ref{fig:pipeline}.

\subsection{Multi-Spectral Scene Reconstruction}

Input to our method is a set of multi-spectral images $\mathcal{I}$, captured with five cameras: an RGB camera, denoted as $C_{\text{RGB}}$, and four multi-spectral cameras $C_R$, $C_G$, $C_{\text{RE}}$, and $C_{\text{NIR}}$, capturing red (R), green (G), red edge (RE), and \nir (NIR) wavelengths, respectively.
Though the G and R bands intersect with the RGB spectrum, their narrow 32 nm capture width boosts spectral accuracy. 
We assume that the image subset, represented by $\mathcal{I} = \{\mathcal{I}_{\text{RGB}}, \mathcal{I}_R, \mathcal{I}_G, \mathcal{I}_{\text{RE}}, \mathcal{I}_{\text{NIR}}\}$, has individual poses and intrinsic parameters for each camera. Additionally, it is assumed that a globally aligned sparse point cloud is initially available, e.g., obtained by the methodology that is presented in Meyer et al.~\cite{meyer2025multi}.


\subsection{Strategies}
In this section, we examine the proposed strategies: \textsc{Separate}, \textsc{Split}, and \textsc{Joint}. A conceptual representation is provided in Fig.\ref{fig:pipeline}.


\subsubsection{\textsc{Separate} Strategy}
The na\"ive baseline for multi-spectral Gaussian splatting is to optimize one individual Gaussian model per spectral modality. 
As such, we split the captured dataset by spectral band (RGB, red, green, red-edge, and \nir) and optimize five individual 3DGS models, without any cross-spectral optimization.
In this context, colors for each Gaussian are formulated conventionally through spherical harmonics, with channel counts varying by model: three for RGB and one for other spectral bands.
\begin{figure}[b]
    \centering
    \includegraphics[width=1\linewidth]{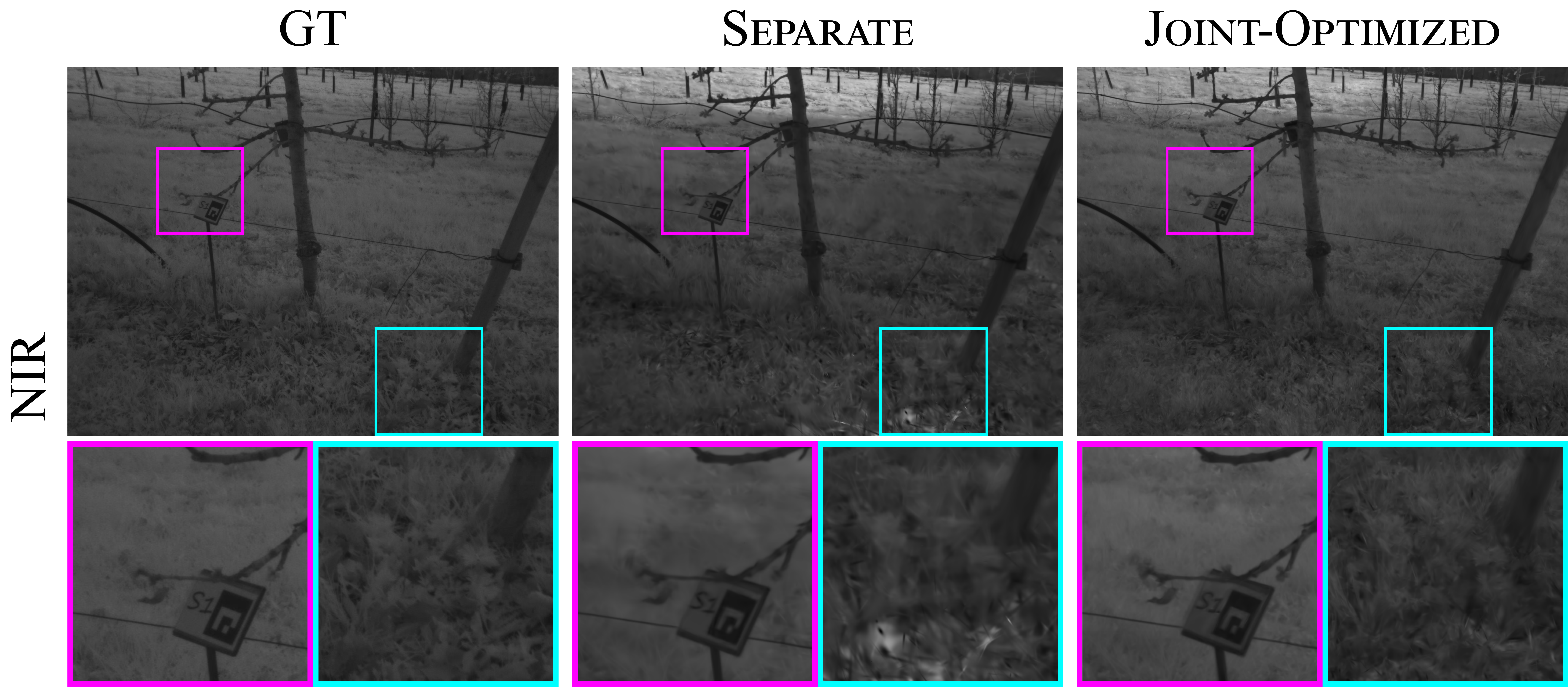}
    \caption{A visual comparison of the NIR reconstructions using the \textsc{Separate} strategy and the \textsc{Joint-Optimized} configuration is shown. The \textsc{Separate} strategy significantly lacks geometric detail, whereas the same region in the jointly optimized reconstruction exhibits improved geometry as a result of the cross-talk effect.}
    \label{fig:motivation-vis-comp}
\end{figure}

A motivation for exploring additional strategies is shown in Fig~\ref{fig:motivation-vis-comp}, as \textsc{Separate} (middle) is shown to have degraded geometric detail compared to the ground truth.




\subsubsection{\textsc{Split} Strategy}
The second strategy follows a staged optimization process. Initially, the model is optimized exclusively using RGB data for the first optimization duration. Afterward, the optimized RGB Gaussian model is \textit{split} into separate models for each available multi-spectral band, which are then optimized independently per spectral channel.

During this splitting process, the geometry, represented by Gaussian primitives reconstructed from the RGB data, is duplicated. The means ($\boldsymbol{\mu}$), covariances ($\boldsymbol{\Sigma}$), and opacities ($\alpha_{i}$) are preserved, while the color component is replaced with the respective spectral band. All parameters of the resulting spectral models are subsequently optimized, and ADC is used for the second optimization cycle.

This strategy capitalizes on RGB-3DGS's strong geometric reconstruction, leveraging the associated ADC heuristics. 
This establishes a solid geometry proxy for refining per-spectra models, enabling unidirectional cross-talk from the RGB domain to other spectral bands, while explicitly preventing reciprocal influence.

\subsubsection{\textsc{Joint} Strategy}
Expanding on the concept of enhancing spectral cross-talk, we examine a third strategy that supports the simultaneous joint optimization of all spectral bands. In this approach, each Gaussian is assigned a series of spherical harmonics parameters for each channel (seven in this instance). Within a training iteration, a channel is randomly selected, and the Gaussian scene is rendered solely in the selected spectral band while disregarding the other spectral color parameters. This approach allows for the joint optimization of both the distinct (spectral) color and the shared Gaussian structural parameters.

\subsubsection{Improving the \textsc{Joint} Strategy}
\label{sec:optims}
The \textsc{joint} strategy allows us further exploration of how to exploit spectral cross-talk, and how to integrate densification in the pipeline.
In Sec.~\ref{sec:ablation}, we ablate the effects of each of these configurations.

\subsubsection*{Spectral Delay (SpecDelay)}
The ADC module lacks spectral data fine-tuning, initially causing spectral gradient heuristics to introduce outliers. 
As such, we investigate deferring the joint optimization during training.
Initially, across the first $n$ iterations (we use 30,000), we limit samples to the RGB pool to ensure robust geometric reconstruction. Subsequently, joint optimization of all spectral channels commences.
In contrast to the \textsc{Split} strategy, with this method, spectral cross-talk is still possible between RGB and spectral channels in both directions, as all bands are continuously optimized after the delayed start. See Fig.~\ref{fig:training_scheme} for a schematic visualization of \textit{SpecDelay}. The objective of this training configuration is to evaluate whether postponing spectral optimization and leveraging the strong geometric reconstruction of RGB data improves the reconstruction quality of the spectral bands.

\subsubsection*{Extended ADC (ExtADC)}
Additionally, we propose an extended ADC configuration, where the densification interval is increased to 60,000 iterations. The objective of this setting is to evaluate whether an extended ADC can further improve reconstruction quality, as the inclusion of multi-spectral images introduces a larger set of training samples compared to standard 3DGS.

\begin{figure}[t]

\resizebox{\linewidth}{!}{%
\begin{tikzpicture}[x=0.075cm, y=1.125cm]
    \fill[gray!25] (-25,  0.5) rectangle (140, -0.7);  
    \fill[gray!10] (-25, -0.7) rectangle (140, -1.9);  
    \fill[gray!25] (-25, -1.9) rectangle (140, -3.1);  
    \fill[gray!10] (-25, -3.1) rectangle (140, -4.4);  
    
  \node[left] at (0,-0.1) {{Default}};
  \draw[thick, black] (0,0) -- (120,0)
    node[midway, above] {\footnotesize all channels};
  \filldraw[black] (120,0) circle (3pt);
  \node[right, xshift=0.2cm] at (120,0) {120,000};

  \draw[thick, red]
    (0,-0.3) -- node[midway, below, black, font=\footnotesize] {ADC} (25,-0.3);
  \filldraw[red] (25,-0.3) circle (3pt);
  \node[right] at (27,-0.3) {25,000};

 \node[left] at (0,-1.4) {{\shortstack{Default\\+ \textit{ExtADC}}}};
  \draw[thick, black] (0,-1.2) -- (120,-1.2)
    node[midway, above] {\footnotesize all channels};
  \filldraw[black] (120,-1.2) circle (3pt);
  \node[right, xshift=0.2cm] at (120,-1.2) {120,000};

  \draw[thick, red]
    (0,-1.5) -- node[midway, below, black, font=\footnotesize] {ADC} (60,-1.5);
  \filldraw[red] (60,-1.5) circle (3pt);
  \node[right] at (62,-1.5) {60,000};

  \node[left] at (0,-2.5) {{\textit{SpecDelay}}};
  \draw[thick, black] (0,-2.4) -- (120,-2.4)
    node[midway, above] {\footnotesize all channels};
  \filldraw[black] (120,-2.4) circle (3pt);
  \node[right, xshift=0.2cm] at (120,-2.4) {120,000};

  \draw[thick, purple] (0,-2.4) -- node[midway, above, text=purple, font=\footnotesize] {RGB only} (30,-2.4);
  \filldraw[purple] (30,-2.4) circle (3pt);
  \node[above right, text=purple] at (30,-2.4) {30,000};

  \draw[thick, red]
    (0,-2.7) -- node[midway, below, black, font=\footnotesize] {ADC} (25,-2.7);
  \filldraw[red] (25,-2.7) circle (3pt);
  \node[right] at (27,-2.7) {25,000};

  \node[left] at (0,-3.8) {{\shortstack{\textit{SpecDelay}\\+ \textit{ExtADC}}}};
  \draw[thick, black] (0,-3.6) -- (120,-3.6)
    node[midway, above] {\footnotesize all channels};
  \filldraw[black] (120,-3.6) circle (3pt);
  \node[right, xshift=0.2cm] at (120,-3.6) {120,000};

  \draw[thick, purple] (0,-3.6) -- node[midway, above, text=purple, font=\footnotesize] {RGB only} (30,-3.6);
  \filldraw[purple] (30,-3.6) circle (3pt);
  \node[above right, text=purple] at (30,-3.6) {30,000};

  \draw[thick, red]
    (0,-3.9) -- node[midway, below, black, font=\footnotesize] {ADC} (60,-3.9);
  \filldraw[red] (60,-3.9) circle (3pt);
  \node[right] at (62,-3.9) {60,000};

\foreach \y in {0, -1.2, -2.4, -3.6} {
    \draw[cyan] (1, \y + 0.1) -- (1, \y - 0.1);
    \node at (1, \y + 0.25) {\includegraphics[width=0.3cm]{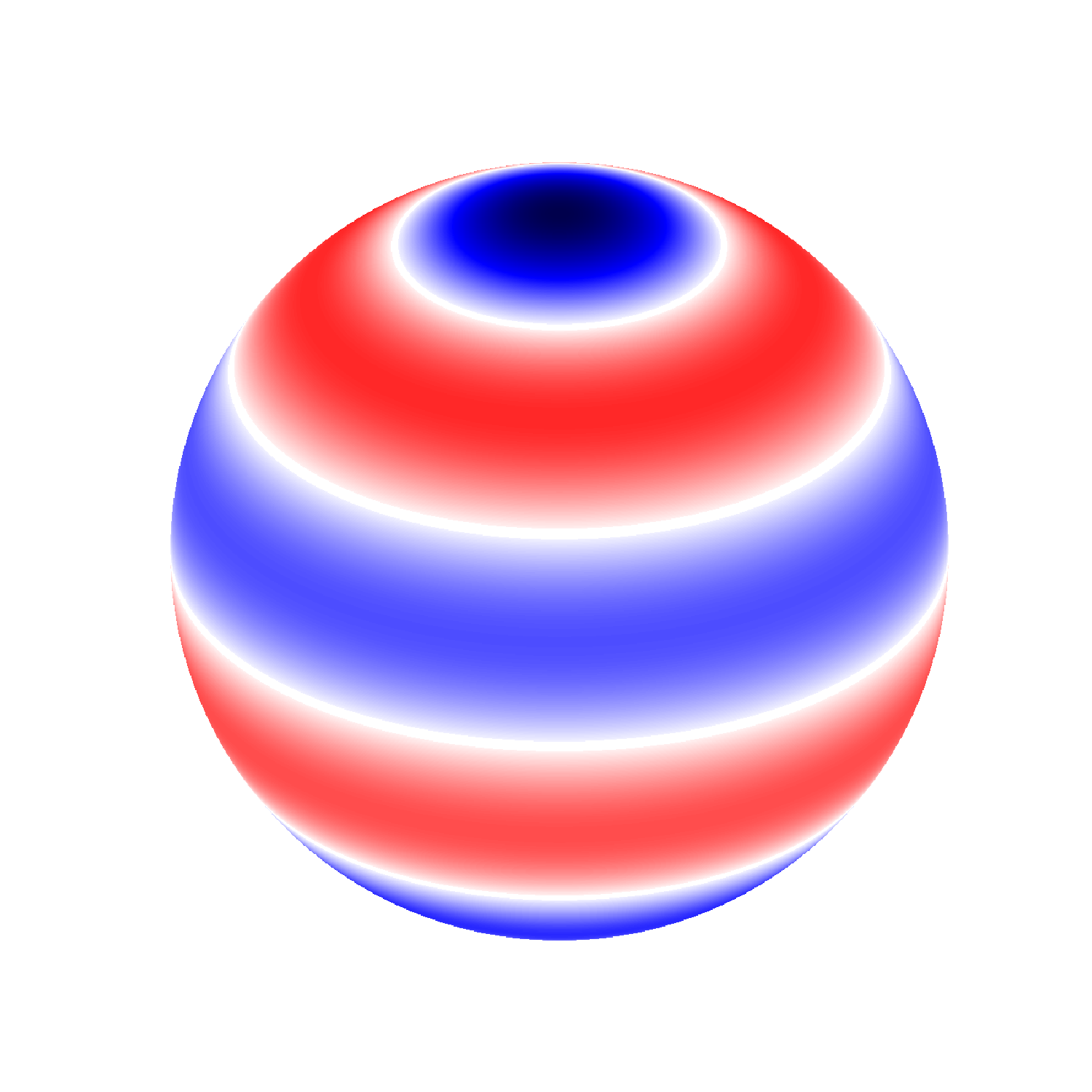}};
}

\end{tikzpicture}
}
    \caption{Training schemes for the improvement modules for the \textsc{joint} strategy. "Extended ADC" (\textit{ExtADC}) extends the densification period, "Spectral Delay" (\textit{SpecDelay}) uses an RGB-only reconstruction during the first 30,000 iterations. The spherical harmonic indicates that for the initialization step of 500 iterations only the color representation is optimized. }
    \label{fig:training_scheme}
\end{figure}

A meaningful combination of the extended ADC is with the \textit{SpecDelay} strategy. In the default \textit{SpecDelay} configuration, the densification steps are typically completed before the spectral channels are incorporated, which may limit their contribution to the reconstruction process. 
Therefore, we additionally investigate whether extending the densification interval to 60,000 iterations can better exploit the additional information provided by the spectral data.
See Fig.~\ref{fig:training_scheme} for a schematic of only \textit{ExtADC} or a combined interaction.

\subsubsection*{Multi-Spectral Aware Densification (MSAD)}
Meyer et al.~\cite{meyer2025multi} introduce a specialized densification heuristic, whereby the per-primitive gradients are tracked independently for each spectral band, and splitting or cloning is done if any of the accumulated gradients exceed the threshold.
This densification strategy always extends to 60,000 iterations.

\subsubsection*{Spectral-Invariant Geometry (SIG)}

Based on the non-RGB inputs, we further explore the behavior when the Gaussian structural parameters (positions $\mu$, covariances $\Sigma$, opacities $\alpha_i$) are held fixed and not updated during optimization.
During every training step, only samples from the RGB bands are allowed to modify geometry, while NIR, red-edge, and other spectra solely refine per-Gaussian spectral coefficients. 
By permanently “locking” shape updates to RGB data, we ensure that low-signal or misaligned spectral channels can not distort the underlying 3D structure; however, this limits spectral cross-talking.

\begin{table*}[t]
\tiny
\centering
\caption[]{\label{tab:strat_comp}Quantitative results of training strategies on the multi-spectral dataset~\cite{meyer2025multi}, averaged over all scenes. The colors indicate the \colorbox{green!40}{best} and \colorbox{red!40}{worst} results.}
\resizebox{\textwidth}{!}{
\begin{tabular}{ l | c c c c c c | c c c c c c | c c c c c c }
\toprule
& \multicolumn{6}{c|}{PSNR $(\uparrow)$} & \multicolumn{6}{c|}{SSIM $(\uparrow)$} & \multicolumn{6}{c}{LPIPS $(\downarrow)$}\\
\midrule
& All & RGB & G & R & RE & NIR & All & RGB & G & R & RE & NIR & All & RGB & G & R & RE & NIR\\
\midrule
\textsc{Separate} & \cred 23.62 & 20.86 & \cred 22.92 & \cred 25.23 & \cred 23.33 & \cred 25.75 & 0.721 & 0.623 & \cred 0.667 & \cred 0.769 & \cred 0.730 & 0.815 & 0.334 & \cg 0.283 & 0.371 & \cred 0.349 & 0.331 & \cred 0.336\\
\textsc{Split} & 24.76 & 20.86 & 24.49 & 26.56 & 25.00 & 26.91 & 0.743 & 0.623 & 0.708 & 0.808 & 0.757 & 0.821 & \cg 0.262 & \cg 0.283 & \cg 0.285 & \cg 0.241 & \cg 0.253 & 0.250\\
\textsc{Joint} & 23.94 & \cred 19.88 & 23.92 & 25.92 & 23.92 & 26.15 & \cred 0.715 & \cred 0.554 & 0.694 & 0.785 & \cred 0.730 & \cred 0.813 & \cred 0.358 & \cred 0.405 & \cred 0.383 & 0.328 & \cred 0.354 & 0.320\\
\textsc{Joint-Optimized} & \cg 24.99 & \cg 21.29 & \cg 24.62 & \cg 26.61 & \cg 25.07 & \cg 27.42 & \cg 0.755 & \cg 0.647 & \cg 0.719 & \cg 0.810 & \cg 0.766 & \cg 0.834 & \cg 0.262 & \cg 0.283 & 0.291 & 0.246 & 0.256 & \cg 0.235\\
\bottomrule
\end{tabular}
}
\end{table*}

\subsection{Implementation Details}

Our model builds upon \textit{gsplat} \cite{ye2025gsplat} and is implemented within the \textit{Nerfstudio} \cite{nerfstudio} framework. 

\subsubsection{Training Strategy}
We trained \textsc{Separate} for 30,000 iterations for each channel and with the same hyperparameters for each channel. \textsc{Split}'s RGB was trained for 30,000 and other bands for another 30,000 iterations; thus, 150,000 iterations in total. \textsc{Joint} was trained once for 120,000 iterations.
We used the Adam optimizer with a learning rate of 0.005. All experiments are conducted on a single NVIDIA A40 GPU. 
In training our model, we adhered to the original 3DGS loss function~\cite{kerbl3Dgaussians}, which specifies:
\begin{equation}
    \mathcal{L}=\sum_{\mathcal{I}_j\sim\mathcal{I}}\sum_{I\sim\mathcal{I}_j}(1-\lambda)\mathcal{L}_1(\hat{I},I) + \lambda \mathcal{L}_{\text{D-SSIM}}(\hat{I},I),
    \label{eq:loss}
\end{equation}
where $I$ represents a randomly selected image from the collection $\mathcal{I}_j$, which consists of all images taken by the $j$-th camera, and $\hat{I}$ indicates the equivalent rendered image. The hyperparameter $\lambda$ is set to 0.2. 
\subsubsection{Initialization}

Following Meyer et al.~\cite{meyer2025multi}, the per-point colors of the sparse point cloud derived from the multi-spectral scene reconstruction contain a mixture of RGB and multi-spectral grayscale values. As a result, directly rendering the Gaussian splats would lead to strong positional gradients.

To address this, at the beginning, the base-band component of the spherical harmonics of all points is randomly initialized, while the higher-order bands are set to zero. During the initial 500 iterations, the optimization process focuses exclusively on learning a coarse set of colors. This procedure is applied consistently across all three training strategies.

\subsubsection{SpecDelay + ExtADC}
For the \textit{SpecDelay} and extended ADC configuration, we introduce minor refinements to ensure a smoother and more stable training process. In the case of \textit{SpecDelay}, the multi-spectral color values are still randomly initialized when introduced, which can potentially distort the geometric reconstruction learned from the RGB data. To mitigate this, we incorporate a multi-spectral color warm-up step.

Specifically, the densification process is paused around the moment multi-spectral bands are included (at 30,000 iterations) for 3,000 iterations between iteration 29,000 and 32,000. During this period, the first 1,000 iterations serve as a short splat refinement step, while the final 2,000 iterations proceed without position, opacity, scale, and orientation optimization, analogous to the initial 500 iterations of the standard initialization step. This gradual introduction stabilizes the integration of the spectral channels and preserves the quality of the RGB-guided geometry.

\section{Evaluation}

We evaluate the strategies on the multi-spectral dataset introduced by Meyer et al.~\cite{meyer2025multi}.
The dataset consists of seven outdoor scenes, recorded with a DJI Mavic 3M drone~\cite{dji_mavic_3m} with RGB, red (R), green (G), red-edge (RE), and \nir (NIR) bands captured.
We also use the poses and initial point clouds provided.
Each scene in the dataset comprises between 81 and 136 images per channel, with a split of 90\% for training and 10\% for validation.
The resolution of the training and rendered RGB images as well as the multi-spectral images is $1800$px$\times1350$px. 

\subsection{Strategy Comparisons}

We compare the strategies \textsc{Separate}, \textsc{Split}, and \textsc{Joint} from Sec.~\ref{sec:method} with the standard image metrics PSNR, SSIM, and LPIPS.
Please note that \textsc{Separate} encompasses individual 3DGS~\cite{kerbl3Dgaussians} reconstructions per band and as such represents the 3DGS baseline as well.
We further compare with an optimized version \textsc{Joint-Optimized}, which is the result of the best combination of modules from Sec.~\ref{sec:optims} (see also Sec.~\ref{sec:ablation} for ablations).

\begin{figure*}
    \centering
    \includegraphics[width=\textwidth]{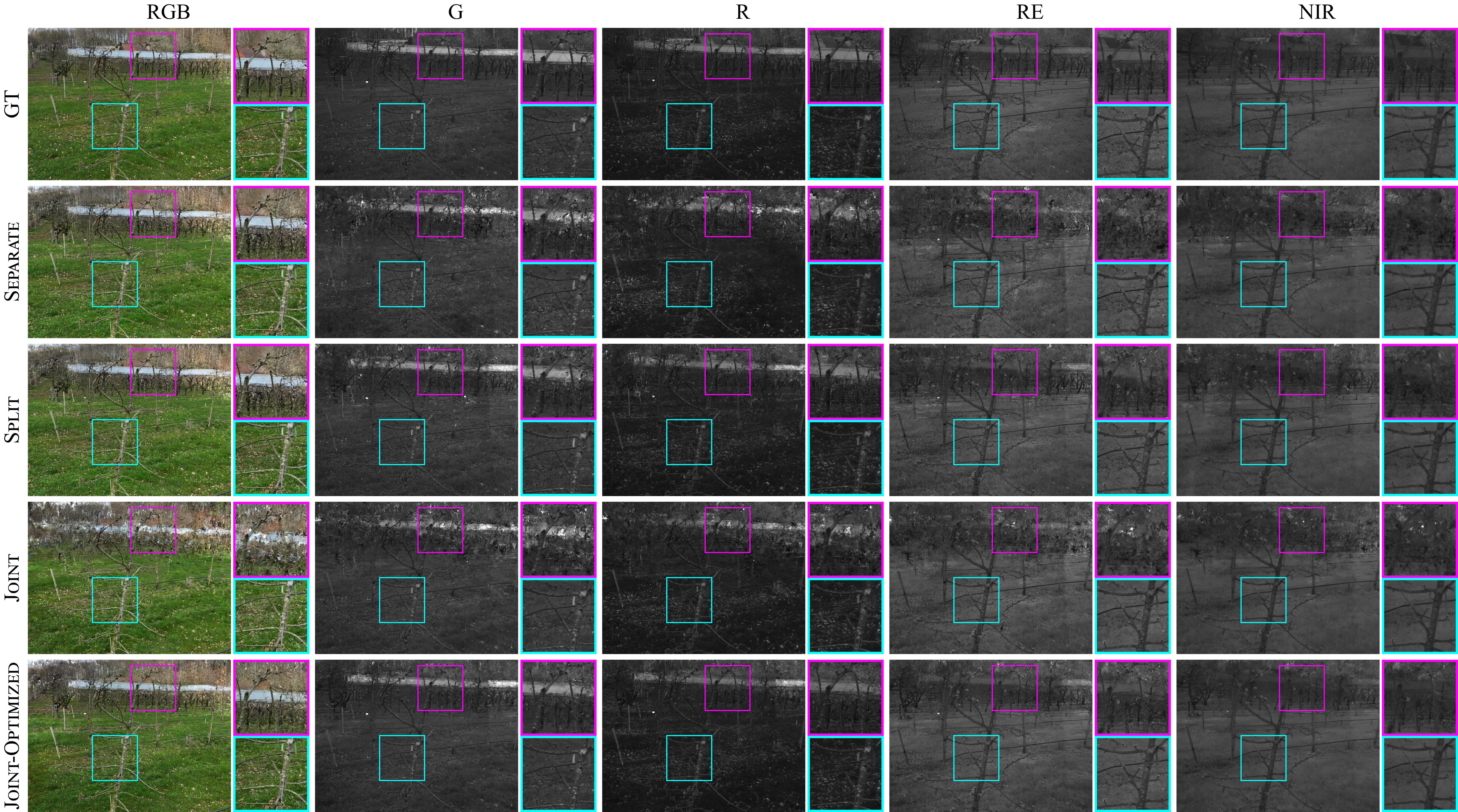}
    \caption{Visual comparison on the \textsc{Single Tree} scene with all available spectral bands.}
    \label{fig:comp_images_ms_single_tree}
\end{figure*}

As shown in Tab.~\ref{tab:strat_comp}, \textsc{Separate} yields the poorest results, particularly for non-RGB reconstruction, indicating that pure 3DGS per channel falls short in multi-spectral scenarios.
\textsc{Split} offers balanced enhancements for non-RGB images. The na\"ive \textsc{Joint} method ranks intermediate in quality, while its optimized version (\textsc{Joint-Optimized}) excels in all renderings, and especially boosts RGB reconstruction by leveraging other spectral bands. A visual comparison of all strategies for one scene is depicted in Fig.~\ref{fig:comp_images_ms_single_tree}. A comprehensive evaluation and visualization of all strategies can be found in the supplements.

\begin{table*}
\tiny
\centering
\caption[]{\label{tab:ablation}Ablations of the optimizations described in Sec.~\ref{sec:optims} average across all datasets. Extended ADC (\textup{ExtADC}) densifies for more iterations, Spectral-Invariant Geometry (\textup{SIG}) only uses RGB samples for geometry updates, Multi-Spectral Aware Densification (\textup{MSAD}) is a spectrally optimized densification strategy~\cite{meyer2025multi}, and Spectral Delay (\textup{SpecDelay}) only uses RGB at the beginning of the optimization. $^*$For this variant, \textup{ExtADC}'s densification interval is decreased (from 60,000) to 45,000 iterations because the total number of iterations is set to 60,000, which otherwise would preclude any fine-tuning of the model. The colors indicate the \colorbox{green!40}{best} and \colorbox{red!40}{worst} results. }

\resizebox{\textwidth}{!}{
\begin{tabular}{ l | c c c c c c | c c c c c c | c c c c c c }
\toprule
& \multicolumn{6}{c|}{PSNR $(\uparrow)$} & \multicolumn{6}{c|}{SSIM $(\uparrow)$} & \multicolumn{6}{c}{LPIPS $(\downarrow)$}\\
\midrule
& All & RGB & G & R & RE & NIR & All & RGB & G & R & RE & NIR & All & RGB & G & R & RE & NIR\\
\midrule
\textsc{Separate} & 23.62 & 20.86 & \cred 22.92 & 25.23 & 23.33 & 25.75 & 0.721 & 0.623 & 0.667 & 0.769 & 0.730 & 0.815 & 0.334 & 0.283 & 0.371 & 0.349 & 0.331 & 0.336\\
\midrule
\textsc{Split} & 24.76 & 20.86 & 24.49 & 26.56 & 25.00 & 26.91 & 0.743 & 0.623 & 0.708 & 0.808 & 0.757 & 0.821 & 0.262 & 0.283 & \cg 0.285 & \cg 0.241 & \cg 0.253 & 0.250\\
\textsc{Split} + \textit{ExtADC}$^*$ & 24.70 & 20.93 & 24.52 & \cg 26.61 & 24.17 & 27.25 & 0.746 & 0.629 & 0.711 & 0.810 & 0.747 & 0.833 & 0.263 & 0.271 & \cg 0.285 & 0.248 & 0.267 & 0.243\\
\midrule
\textsc{Joint} + \textit{SIG} & \cred 23.11 & \cred 19.57 & 22.98 & \cred 24.76 & \cred 23.12 & \cred 25.19 & \cred 0.673 & \cred 0.515 & \cred 0.650 & \cred 0.746 & \cred 0.683 & \cred 0.774 & \cred 0.380 & \cred 0.407 & \cred 0.408 & \cred 0.361 & \cred 0.379 & \cred 0.348\\
\textsc{Joint} + \textit{SIG} + \textit{ExtADC} & 23.35 & 19.83 & 23.24 & 25.07 & 23.28 & 25.38 & 0.680 & 0.530 & 0.654 & 0.752 & 0.690 & 0.777 & 0.353 & 0.375 & 0.380 & 0.334 & 0.352 & 0.324\\
\textsc{Joint} + \textit{SIG} + \textit{MSAD} & 23.54 & 19.98 & 23.38 & 25.27 & 23.45 & 25.67 & 0.681 & 0.545 & 0.652 & 0.750 & 0.688 & \cred 0.774 & 0.315 & 0.335 & 0.339 & 0.300 & 0.312 & 0.291\\
\textsc{Joint} + \textit{SIG} + \textit{SpecDelay} & 24.21 & 21.09 & 23.81 & 25.81 & 24.09 & 26.34 & 0.725 & 0.631 & 0.681 & 0.783 & 0.730 & 0.802 & 0.272 & 0.272 & 0.303 & 0.262 & 0.270 & 0.253\\
\textsc{Joint} + \textit{SIG} + \textit{SpecDelay} + \textit{ExtADC} & 24.27 & 21.18 & 23.90 & 25.83 & 24.13 & 26.38 & 0.727 & 0.631 & 0.682 & 0.784 & 0.732 & 0.805 & 0.272 & 0.274 & 0.303 & 0.262 & 0.269 & 0.251\\
\textsc{Joint} + \textit{SIG} + \textit{SpecDelay} + \textit{MSAD} & 24.41 & \cg 21.40 & 23.94 & 25.93 & 24.28 & 26.57 & 0.734 & 0.644 & 0.688 & 0.789 & 0.741 & 0.811 & \cg 0.258 & \cg 0.261 & 0.288 & 0.247 & 0.254 & 0.236\\
\textsc{Joint} & 23.94 & 19.88 & 23.92 & 25.92 & 23.92 & 26.15 & 0.715 & 0.554 & 0.694 & 0.785 & 0.730 & 0.813 & 0.358 & 0.405 & 0.383 & 0.328 & 0.354 & 0.320\\
\textsc{Joint} + \textit{ExtADC} & 24.21 & 20.20 & 24.11 & 26.15 & 24.17 & 26.52 & 0.719 & 0.570 & 0.695 & 0.788 & 0.733 & 0.813 & 0.337 & 0.377 & 0.362 & 0.310 & 0.332 & 0.302\\
\textsc{Joint} + \textit{MSAD} & 24.48 & 20.57 & 24.31 & 26.40 & 24.41 & 26.79 & 0.736 & 0.599 & 0.708 & 0.801 & 0.748 & 0.824 & 0.300 & 0.337 & 0.324 & 0.273 & 0.295 & 0.268\\
\textsc{Joint} + \textit{SpecDelay} & 24.80 & 21.03 & \cg 24.62 & 26.56 & 24.83 & 27.04 & 0.747 & 0.634 & 0.713 & 0.807 & 0.757 & 0.827 & 0.273 & 0.292 & 0.301 & 0.256 & 0.268 & 0.247\\
\textsc{Joint} + \textit{SpecDelay} + \textit{ExtADC} & 24.90 & 21.27 & 24.60 & 26.56 & 24.96 & 27.19 & 0.753 & 0.639 & 0.718 & \cg 0.811 & 0.764 & \cg 0.834 & 0.273 & 0.295 & 0.303 & 0.255 & 0.267 & 0.244\\
\textsc{Joint} + \textit{SpecDelay} + \textit{MSAD} & \cg 24.99 & 21.29 & \cg 24.62 & \cg 26.61 & \cg 25.07 & \cg 27.42 & \cg 0.755 & \cg 0.647 & \cg 0.719 & 0.810 & \cg 0.766 & \cg 0.834 & 0.262 & 0.283 & 0.291 & 0.246 & 0.256 & \cg 0.235\\
\bottomrule
\end{tabular}
}
\end{table*}

\subsection{Investigation of \textsc{Joint} Optimizations}
\label{sec:ablation}

We ablate the augmentation components (Sec.~\ref{sec:optims}) Spectral Delay (\textit{SpecDelay}), Extended ADC (\textit{ExtADC}), Multi-Spectral Aware Densification (\textit{MSAD}), and Spectral-Invariant Geometry (\textit{SIG}) in Tab.~\ref{tab:ablation}. We can see that adding \textit{SpecDelay} to \textsc{Joint} increases quality drastically and surpasses \textsc{split}, showcasing the effectiveness of cross-spectral optimization.
\textit{ExtADC} further increases quality, outperforming both base \textsc{Split} as well as \textsc{Split} with \textit{ExtADC} in PSNR and SSIM except in the red and red-edge channels' PSNR.
Adapting \textit{MSAD}~\cite{meyer2025multi} further increases quality across all metrics except the red channel's SSIM.
We label the best-performing \textsc{Joint} + \textit{SpecDelay} + \textit{MSAD} as \textsc{Joint-Optimized} in Tab.~\ref{tab:strat_comp}.

An interesting combination is the \textsc{Joint} \textit{+ SIG + SpecDelay + MSAD} version.
While \textit{SIG} in general yields unsatisfactory results, especially in the non-RGB reconstruction, this variation provides the best overall RGB results in PSNR and LPIPS.
In this version, after an initial RGB-only optimization, spectral channels only cross-talk densification information to RGB 
(via \textit{MSAD}), enabling materials with spectral variance to improve geometry detail and subsequently enhance RGB renderings.
In contrast to the \textsc{Split} strategy, this shows the benefits of spectral cross-talk for geometry.
Exploring this direction further may be a great option for future improvements of RGB renderings through spectral captures.

\subsection{Cross-Talking Investigation}
\begin{table}
\tiny
\caption[]{Spectral cross-talking effects with RGB, green (G), red (R), red-edge (RE) and \nir (NIR) in the \textsc{Joint-Optimized} configuration. The colors indicate the \colorbox{green!40}{best} and \colorbox{red!40}{worst} results.}
\label{tab:cross_talk_investigation}
\setlength\tabcolsep{2pt}
\resizebox{1.02\columnwidth}{!}{
\begin{tabular}{ l | c c c c c | c c c c c | c c c c c }
\toprule
& \multicolumn{5}{c|}{PSNR $(\uparrow)$} & \multicolumn{5}{c|}{SSIM $(\uparrow)$} & \multicolumn{5}{c}{LPIPS $(\downarrow)$}\\
\midrule
& RGB & G & R & RE & NIR & RGB & G & R & RE & NIR & RGB & G & R & RE & NIR\\
\midrule
RGB & \cred 21.06 & - & - & - & - & \cred 0.637 & - & - & - & - & 0.271 & - & - & - & -\\
RGB + G & 21.16 & 24.33 & - & - & - & 0.641 & 0.711 & - & - & - & 0.275 & \cg 0.284 & - & - & -\\
RGB + R & 21.22 & - & \cred 26.30 & - & - & 0.644 & - & \cred 0.806 & - & - & 0.271 & - & 0.241 & - & -\\
RGB + RE & 21.26 & - & - & 24.77 & - & 0.645 & - & - & 0.758 & - & \cg 0.269 & - & - & \cg 0.249 & -\\
RGB + NIR & 21.19 & - & - & - & 27.15 & 0.645 & - & - & - & 0.828 & 0.270 & - & - & - & \cg 0.232\\
RGB + G + R & 21.20 & 24.45 & 26.51 & - & - & 0.642 & 0.712 & 0.807 & - & - & 0.278 & 0.289 & 0.245 & - & -\\
RGB + G + RE & 21.07 & \cred 24.28 & - & \cred 24.69 & - & 0.641 & \cred 0.710 & - & \cred 0.755 & - & 0.281 & 0.291 & - & 0.260 & -\\
RGB + G + NIR & 21.16 & 24.39 & - & - & \cred 27.00 & 0.639 & \cred 0.710 & - & - & \cred 0.821 & \cred 0.285 & \cred 0.294 & - & - & \cred 0.245\\
RGB + R + RE & 21.17 & - & 26.45 & 24.85 & - & 0.644 & - & 0.808 & 0.759 & - & 0.282 & - & \cred 0.248 & \cred 0.261 & -\\
RGB + R + NIR & 21.15 & - & 26.62 & - & 27.13 & 0.645 & - & 0.808 & - & 0.825 & 0.279 & - & 0.246 & - & 0.237\\
RGB + RE + NIR & 21.16 & - & - & 24.82 & 27.15 & 0.642 & - & - & 0.758 & 0.827 & 0.280 & - & - & 0.257 & 0.237\\
RGB + G + R + RE & \cg 21.40 & \cg 24.63 & \cg 26.66 & 25.02 & - & \cg 0.649 & \cg 0.719 & \cg 0.813 & \cg 0.766 & - & 0.276 & \cg 0.284 & \cg 0.240 & 0.251 & -\\
RGB + G + R + NIR & 21.24 & 24.59 & 26.65 & - & 27.25 & 0.645 & 0.718 & 0.812 & - & 0.831 & 0.282 & 0.289 & 0.244 & - & 0.236\\
RGB + G + RE + NIR & 21.22 & 24.30 & - & 24.81 & 27.10 & 0.644 & 0.714 & - & 0.762 & 0.829 & 0.283 & 0.292 & - & 0.258 & 0.238\\
RGB + R + RE + NIR & 21.30 & - & 26.53 & 25.00 & 27.30 & 0.648 & - & 0.810 & 0.765 & 0.832 & 0.281 & - & 0.245 & 0.255 & 0.234\\
\text{All Bands} & 21.29 & 24.62 & 26.61 & \cg 25.07 & \cg 27.42 & 0.647 & \cg 0.719 & 0.810 & \cg 0.766 & \cg 0.834 & 0.283 & 0.291 & 0.246 & 0.256 & 0.235\\
\bottomrule
\end{tabular}
}
\end{table}

To further investigate spectral cross-talk (beyond SIG), we conducted an additional series of experiments. As the underlying configuration, we consistently employed our \textsc{Joint-Optimized} strategy with 120,000 training iterations. In these experiments, we systematically extended the set of RGB images by successively adding different spectral bands in all possible permutations. The goal of this evaluation is to quantify the impact of spectral cross-talk by measuring the reconstruction quality for each individual band.
Tab.~\ref{tab:cross_talk_investigation} summarizes the results of this analysis averaged over all datasets. Starting from a plain RGB-only reconstruction, which yields the lowest PSNR and SSIM values, the successive inclusion of additional spectral bands leads to a steady improvement in reconstruction quality. Specifically, extending from RGB to all available bands results in a PSNR improvement of approximately 0.2dB for the RGB channels. The largest gain for RGB is observed when combining RGB with the green, red, and red-edge bands, yielding an improvement of over 0.3dB in PSNR.
An important observation is that spectral cross-talk enhances not only the RGB channels but also the additional spectral channels. Including all available bands, as opposed to dual-band setups like RGB + green, results in nearly a 0.3dB increase in PSNR for each multi-spectral channel. This improvement is due to the shared geometric representation in the 3D Gaussian Splatting model, where all spectral bands, to varying extents, contribute to the scene's geometry. A detailed per-scene evaluation can be found in the supplementary material.

\subsection{Number of Primitives}

Additionally, we investigated the number of Gaussian splats required to represent a scene for all evaluated strategies. For the \textsc{Garden} scene, we found that the \textsc{Separate} variant requires, on average, one million splats per spectral band, while our best-performing \textsc{Joint-Optimized} approach requires only two million primitives to represent the same scene with increased quality in a joint representation. Furthermore, we observed that the multi-spectral aware densification strategy (MSAD) increases the number of splats by approximately $100\%$ compared to the plain \textsc{Joint} variant. This highlights the selective capability of MSAD to increase the number of scene primitives, specifically in spectral bands where higher detail is required. An overview of the number of primitives for two scenes is provided in the supplementary material.




\section{Limitations}


In the preceding section, we demonstrated the effect of the \textsc{Joint-Optimized} configuration on the quantitative improvements, which can be attributed to spectral cross-talk.
A major limitation, however, is that only geometric cross-talk is possible, as the \textsc{Joint-Optimized} configuration allows for joint optimization of geometric properties but does not affect the spectral color representation.
In this context, recent neural approaches~\cite{meyer2025multi} have shown promising reconstruction results by jointly optimizing color values through per-splat feature vectors in combination with a shared MLP.

Another limitation of the \textsc{Joint} approach is the assumption of a shared scene geometry across all spectral bands. Although this is appropriate for the visible and \nir spectra examined in our study, it may introduce reconstruction errors when applied to thermal infra red imagery. This is because the observed geometry can differ significantly between RGB and thermal bands, due to differences in how objects emit or reflect radiation at those wavelengths.

\section{Conclusion \& Future Work}

To conclude, we presented a comprehensive study of training strategies for multi-spectral scene reconstruction based on 3D Gaussian Splatting (3DGS). We demonstrated that a joint optimization of RGB and multi-spectral data, combined with moderate adjustments to the training configuration, can significantly enhance reconstruction quality, both visually and according to quantitative metrics. Furthermore, we highlighted the effect of spectral cross-talk, where the interplay between RGB and multi-spectral data improves reconstruction through geometric and color-based information transfer.


In particular, the field of plant phenotyping, both in agricultural environments and greenhouse settings, would greatly benefit from a multi-spectral 3DGS approach. Beyond vegetation index computation, as demonstrated by Meyer et al.~\cite{meyer2025multi},
our method shows potential for applications such as fruit counting~\cite{FruitNeRF, FruitNeRFpp}. Moreover, real-time implementations such as LiveNVS~\cite{livenvs} or immersive multi-spectral Gaussian Splatting systems~\cite{franke2025vrsplatting} could make these capabilities far more accessible for practical agricultural use.

\section*{Acknowledgments}
Lukas Meyer was funded by the 5G innovation program of the German Federal Ministry for Digital and Transport under the funding code 165GU103C.
Maximilian Weiherer was funded by the German Federal Ministry of Education and Research (BMBF), FKZ: 01IS22082 (IRRW). 
The authors are responsible for the content of this publication.
The authors thank the reviewers for their insightful comments.
The authors gratefully acknowledge the scientific support and HPC resources provided by the Erlangen National High Performance Computing Center (NHR@FAU) of the Friedrich-Alexander-Universität Erlangen-Nürnberg (FAU) under the NHR project \textit{b162dc}. NHR funding is provided by federal and Bavarian state authorities. NHR@FAU hardware is partially funded by the German Research Foundation (DFG) – 440719683.

\bibliographystyle{eg-alpha} 
\bibliography{bib}

\beginsupplement

\setcounter{section}{0}


\begin{figure*}[h!]
    \center
    \includegraphics[width=0.95\textwidth]{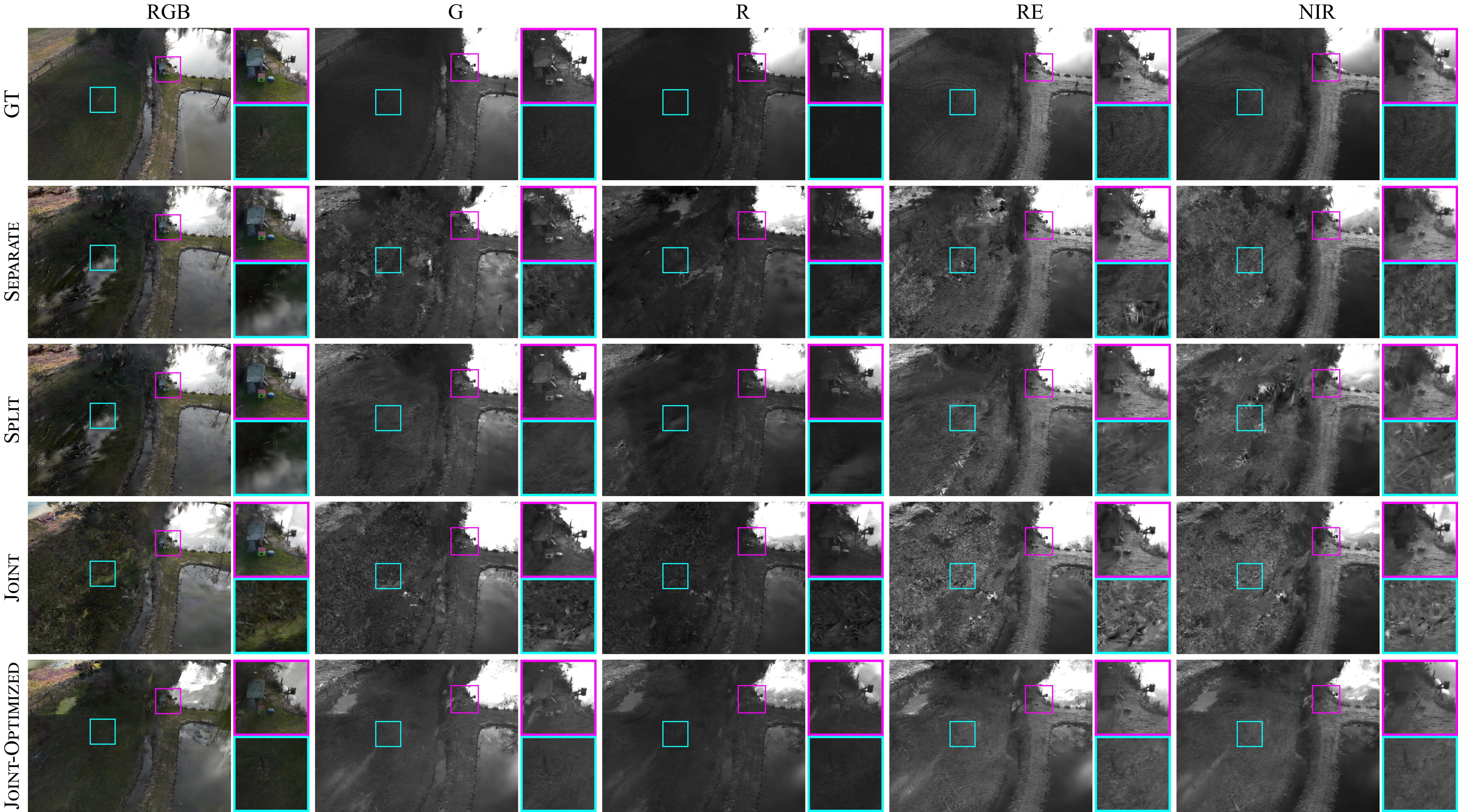}

    \caption{Visual comparison on the \textsc{Lake} scene with all available spectral bands.}
    \label{fig:comp_images_lake}
\end{figure*}

\begin{table*}[h!]

\caption[]{Comparison on \textsc{Lake} scene. The colors indicate the \colorbox{green!40}{best} and \colorbox{red!40}{worst} results.}

\tiny
\centering

\resizebox{\textwidth}{!}{
\begin{tabular}{ l | c c c c c c | c c c c c c | c c c c c c }
\toprule
& \multicolumn{6}{c|}{PSNR $(\uparrow)$} & \multicolumn{6}{c|}{SSIM $(\uparrow)$} & \multicolumn{6}{c}{LPIPS $(\downarrow)$}\\
\midrule
& All & RGB & G & R & RE & NIR & All & RGB & G & R & RE & NIR & All & RGB & G & R & RE & NIR\\
\midrule
\textsc{Separate} & 21.16 & 20.76 & 20.88 & 24.63 & 19.54 & 20.01 & 0.658 & 0.587 & \cred 0.616 & 0.764 & 0.627 & 0.696 & 0.459 & 0.500 & 0.518 & \cred 0.467 & \cg 0.407 & 0.402\\
\textsc{Split} & 21.12 & 20.76 & \cg 22.37 & 24.62 & 19.57 & \cred 18.28 & 0.642 & 0.587 & 0.645 & 0.766 & 0.599 & \cred 0.613 & 0.456 & 0.500 & 0.458 & 0.401 & 0.437 & \cred 0.484\\
\textsc{Split} + \textit{ExtADC}$^*$ & 20.86 & \cred 20.09 & 21.86 & 24.71 & \cred 17.67 & 19.99 & 0.641 & 0.572 & 0.640 & 0.765 & \cred 0.536 & 0.690 & 0.438 & 0.496 & 0.451 & 0.401 & 0.476 & \cg 0.366\\
\textsc{Joint} + \textit{SIG} & 20.64 & 20.37 & 21.26 & 24.21 & 18.45 & 18.93 & 0.656 & 0.593 & 0.657 & 0.770 & 0.588 & 0.669 & \cred 0.507 & \cred 0.546 & \cred 0.533 & 0.463 & \cred 0.520 & 0.471\\
\textsc{Joint} + \textit{SIG} + \textit{ExtADC} & 21.12 & 20.68 & 21.80 & 24.53 & 18.92 & 19.68 & 0.652 & 0.587 & 0.653 & 0.768 & 0.584 & 0.666 & 0.472 & 0.517 & 0.497 & 0.431 & 0.481 & 0.435\\
\textsc{Joint} + \textit{SIG} + \textit{MSAD} & 21.29 & 20.64 & 21.88 & 24.90 & 19.14 & 19.87 & \cred 0.634 & \cred 0.567 & 0.634 & 0.752 & 0.571 & 0.647 & 0.452 & 0.493 & 0.475 & 0.416 & 0.455 & 0.420\\
\textsc{Joint} + \textit{SIG} + \textit{SpecDelay} & 21.23 & 21.02 & 21.68 & 24.45 & 19.26 & 19.76 & 0.652 & 0.591 & 0.653 & 0.766 & 0.589 & 0.663 & 0.451 & 0.480 & 0.470 & 0.409 & 0.463 & 0.431\\
\textsc{Joint} + \textit{SIG} + \textit{SpecDelay} + \textit{ExtADC} & 21.58 & 21.34 & 22.19 & 24.70 & 19.63 & 20.03 & 0.657 & 0.595 & 0.654 & 0.767 & 0.596 & 0.671 & 0.441 & 0.472 & 0.465 & 0.403 & 0.451 & 0.414\\
\textsc{Joint} + \textit{SIG} + \textit{SpecDelay} + \textit{MSAD} & 21.32 & 21.33 & 21.71 & 24.32 & 19.35 & 19.88 & 0.661 & 0.606 & 0.651 & 0.764 & 0.606 & 0.678 & \cg 0.409 & \cg 0.445 & \cg 0.437 & \cg 0.374 & 0.414 & 0.374\\
\textsc{Joint} & 21.50 & 21.13 & 22.05 & 24.90 & 19.42 & 20.00 & 0.675 & 0.607 & 0.667 & \cg 0.783 & 0.620 & 0.699 & 0.465 & 0.521 & 0.494 & 0.417 & 0.469 & 0.421\\
\textsc{Joint} + \textit{ExtADC} & 21.46 & 21.10 & 22.03 & 25.17 & 19.14 & 19.84 & 0.666 & 0.603 & 0.657 & 0.775 & 0.610 & 0.686 & 0.450 & 0.504 & 0.474 & 0.411 & 0.452 & 0.408\\
\textsc{Joint} + \textit{MSAD} & \cg 21.85 & 21.16 & \cg 22.37 & \cg 25.38 & 19.74 & \cg 20.61 & 0.673 & 0.605 & 0.661 & 0.779 & 0.622 & 0.698 & 0.422 & 0.482 & 0.448 & 0.387 & 0.418 & 0.375\\
\textsc{Joint} + \textit{SpecDelay} & 20.93 & 20.60 & 21.55 & 24.04 & 18.90 & 19.54 & 0.655 & 0.597 & 0.654 & 0.763 & 0.592 & 0.668 & 0.464 & 0.509 & 0.477 & 0.422 & 0.475 & 0.436\\
\textsc{Joint} + \textit{SpecDelay} + \textit{ExtADC} & 21.66 & \cg 21.49 & 22.06 & 24.56 & \cg 19.85 & 20.31 & \cg 0.679 & \cg 0.620 & \cg 0.668 & 0.778 & \cg 0.628 & \cg 0.703 & 0.434 & 0.484 & 0.461 & 0.396 & 0.437 & 0.392\\
\textsc{Joint} + \textit{SpecDelay} + \textit{MSAD} & \cred 20.13 & 20.38 & \cred 20.23 & \cred 21.95 & 18.68 & 19.42 & 0.651 & 0.603 & 0.641 & \cred 0.729 & 0.604 & 0.676 & 0.445 & 0.493 & 0.470 & 0.418 & 0.443 & 0.402\\
\bottomrule
\end{tabular}
}

\end{table*}

\clearpage

\begin{figure*}[h!]
    \centering
    \includegraphics[width=0.95\textwidth]{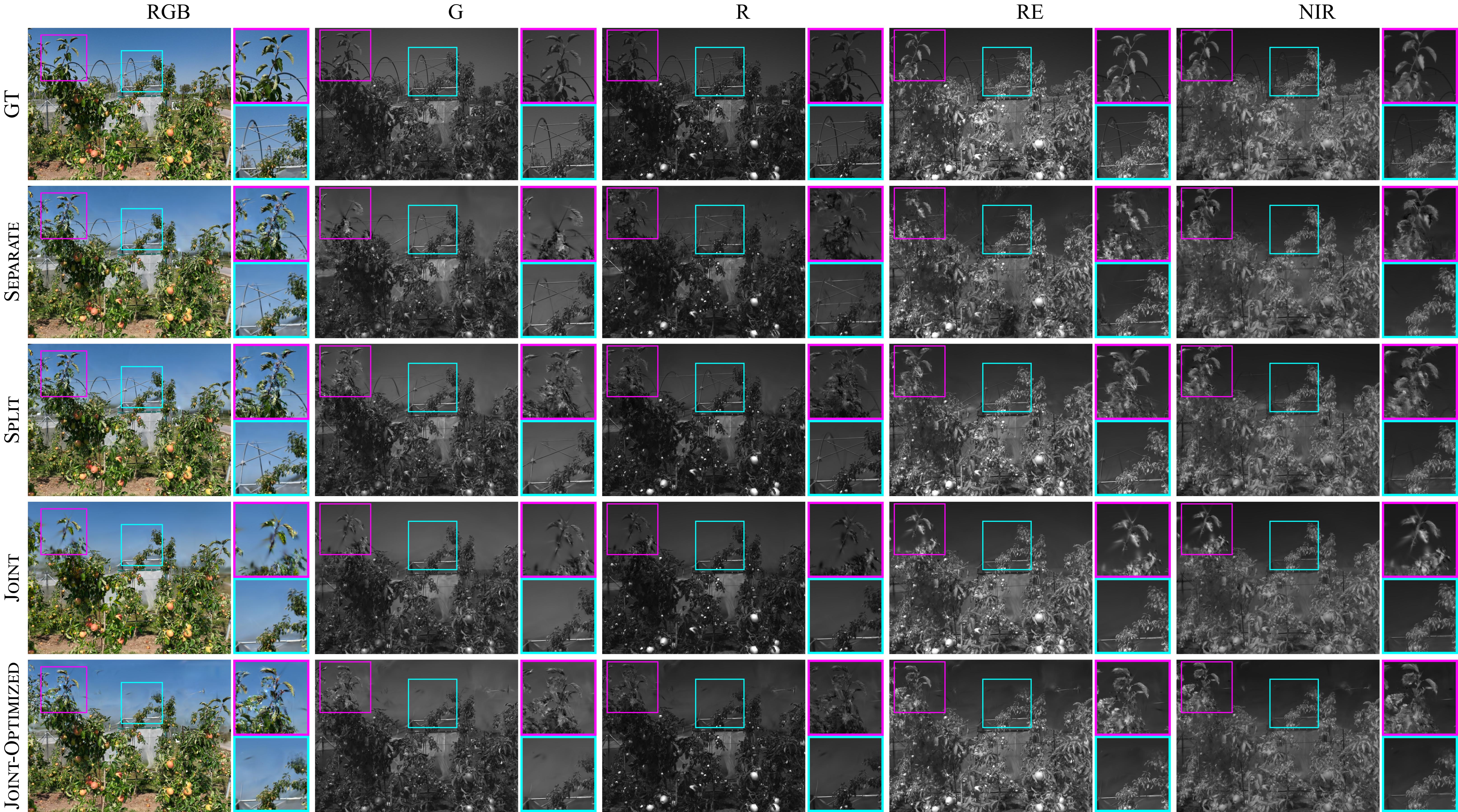}

    \caption{Visual comparison on the \textsc{Fruit Trees} scene with all available spectral bands.}
    \label{fig:comp_images_fruit_trees}
\end{figure*}

\begin{table*}[h!]

\caption[]{Comparison on \textsc{Fruit Trees} scene. The colors indicate the \colorbox{green!40}{best} and \colorbox{red!40}{worst} results.}

\tiny
\centering

\resizebox{\textwidth}{!}{
\begin{tabular}{ l | c c c c c c | c c c c c c | c c c c c c }
\toprule
& \multicolumn{6}{c|}{PSNR $(\uparrow)$} & \multicolumn{6}{c|}{SSIM $(\uparrow)$} & \multicolumn{6}{c}{LPIPS $(\downarrow)$}\\
\midrule
& All & RGB & G & R & RE & NIR & All & RGB & G & R & RE & NIR & All & RGB & G & R & RE & NIR\\
\midrule
\textsc{Separate} & 20.92 & 16.69 & \cred 21.02 & \cred 22.89 & 20.37 & 23.61 & 0.603 & 0.487 & \cred 0.535 & 0.654 & 0.603 & 0.739 & 0.323 & 0.280 & 0.354 & 0.351 & 0.311 & 0.321\\
\textsc{Split} & 21.66 & 16.69 & 21.98 & 24.35 & 21.14 & 24.13 & 0.642 & 0.487 & 0.592 & 0.723 & 0.646 & 0.763 & \cg 0.265 & 0.280 & 0.291 & \cg 0.249 & \cg 0.259 & 0.245\\
\textsc{Split} + \textit{ExtADC}$^*$ & 21.88 & 16.84 & 22.20 & 24.75 & 21.15 & \cg 24.47 & 0.644 & 0.491 & 0.593 & 0.726 & 0.645 & 0.765 & \cg 0.265 & 0.270 & \cg 0.289 & 0.252 & 0.260 & 0.253\\
\textsc{Joint} + \textit{SIG} & \cred 20.52 & \cred 15.41 & 21.42 & 23.10 & \cred 19.97 & \cred 22.73 & \cred 0.559 & \cred 0.359 & \cred 0.535 & \cred 0.650 & \cred 0.557 & \cred 0.695 & \cred 0.375 & \cred 0.388 & \cred 0.404 & \cred 0.367 & \cred 0.369 & \cred 0.345\\
\textsc{Joint} + \textit{SIG} + \textit{ExtADC} & 20.93 & 15.80 & 21.61 & 23.44 & 20.41 & 23.36 & 0.575 & 0.389 & 0.545 & 0.659 & 0.575 & 0.708 & 0.346 & 0.356 & 0.376 & 0.340 & 0.342 & 0.316\\
\textsc{Joint} + \textit{SIG} + \textit{MSAD} & 21.06 & 15.80 & 21.74 & 23.76 & 20.49 & 23.50 & 0.580 & 0.400 & 0.547 & 0.666 & 0.579 & 0.708 & 0.318 & 0.331 & 0.344 & 0.309 & 0.314 & 0.291\\
\textsc{Joint} + \textit{SIG} + \textit{SpecDelay} & 21.62 & 16.94 & 21.98 & 24.24 & 20.97 & 23.98 & 0.631 & 0.496 & 0.577 & 0.706 & 0.631 & 0.743 & 0.274 & 0.271 & 0.309 & 0.266 & 0.271 & 0.252\\
\textsc{Joint} + \textit{SIG} + \textit{SpecDelay} + \textit{ExtADC} & 21.66 & 17.04 & 22.04 & 24.30 & 20.98 & 23.96 & 0.634 & 0.500 & 0.581 & 0.709 & 0.635 & 0.747 & 0.274 & 0.274 & 0.309 & 0.267 & 0.271 & 0.250\\
\textsc{Joint} + \textit{SIG} + \textit{SpecDelay} + \textit{MSAD} & 21.74 & 17.06 & 22.07 & 24.37 & 21.10 & 24.09 & 0.637 & 0.505 & 0.583 & 0.711 & 0.639 & 0.749 & 0.266 & \cg 0.268 & 0.298 & 0.257 & 0.263 & 0.243\\
\textsc{Joint} & 21.70 & 16.38 & 22.29 & 24.60 & 21.03 & 24.23 & 0.620 & 0.430 & 0.587 & 0.708 & 0.623 & 0.751 & 0.346 & 0.382 & 0.375 & 0.330 & 0.335 & 0.306\\
\textsc{Joint} + \textit{ExtADC} & 21.54 & 16.43 & 22.02 & 24.36 & 20.85 & 24.03 & 0.624 & 0.446 & 0.588 & 0.709 & 0.628 & 0.751 & 0.323 & 0.355 & 0.352 & 0.308 & 0.313 & 0.287\\
\textsc{Joint} + \textit{MSAD} & 21.82 & 16.65 & 22.27 & \cg 24.82 & 21.09 & 24.27 & 0.636 & 0.467 & 0.596 & 0.719 & 0.640 & 0.757 & 0.293 & 0.326 & 0.320 & 0.277 & 0.282 & 0.260\\
\textsc{Joint} + \textit{SpecDelay} & 21.92 & 17.04 & 22.29 & 24.65 & \cg 21.29 & 24.35 & 0.653 & 0.502 & 0.605 & 0.733 & 0.657 & 0.769 & 0.272 & 0.286 & 0.304 & 0.259 & 0.265 & 0.245\\
\textsc{Joint} + \textit{SpecDelay} + \textit{ExtADC} & \cg 21.94 & \cg 17.22 & \cg 22.30 & 24.69 & 21.21 & 24.29 & 0.659 & 0.508 & \cg 0.611 & \cg 0.737 & 0.663 & \cg 0.774 & 0.277 & 0.294 & 0.310 & 0.263 & 0.270 & 0.247\\
\textsc{Joint} + \textit{SpecDelay} + \textit{MSAD} & 21.92 & 17.11 & 22.25 & 24.65 & 21.25 & 24.35 & \cg 0.660 & \cg 0.511 & \cg 0.611 & \cg 0.737 & \cg 0.664 & \cg 0.774 & 0.268 & 0.286 & 0.298 & 0.254 & 0.260 & \cg 0.239\\
\bottomrule
\end{tabular}
}

\end{table*}

\clearpage

\begin{figure*}[h!]
    \centering
        \includegraphics[width=\textwidth]{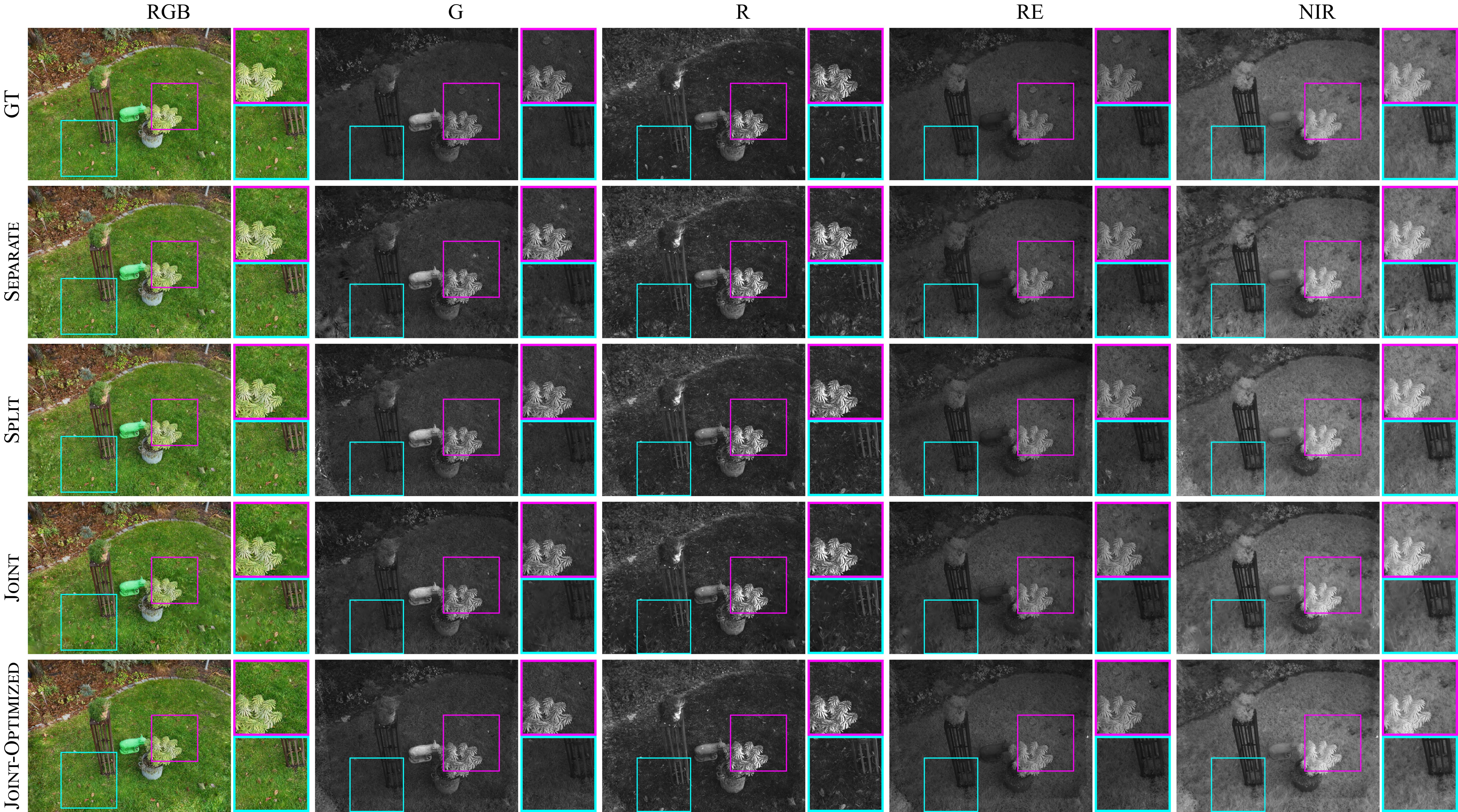}

    \caption{Visual comparison on the \textsc{Garden} scene with all available spectral bands.}
    \label{fig:comp_images_garden}
\end{figure*}

\begin{table*}[h!]

\caption[]{Comparison on \textsc{Garden} scene. The colors indicate the \colorbox{green!40}{best} and \colorbox{red!40}{worst} results.}

\tiny
\centering

\resizebox{\textwidth}{!}{
\begin{tabular}{ l | c c c c c c | c c c c c c | c c c c c c}
\toprule
& \multicolumn{6}{c|}{PSNR $(\uparrow)$} & \multicolumn{6}{c|}{SSIM $(\uparrow)$} & \multicolumn{6}{c}{LPIPS $(\downarrow)$}\\
\midrule
& All & RGB & G & R & RE & NIR & All & RGB & G & R & RE & NIR & All & RGB & G & R & RE & NIR\\
\midrule
\textsc{Separate} & 26.91 & 23.47 & 26.85 & 26.19 & 29.38 & 28.69 & 0.777 & 0.692 & 0.747 & 0.775 & 0.825 & 0.843 & 0.270 & 0.217 & 0.281 & 0.235 & 0.313 & 0.303\\
\textsc{Split} & 27.24 & 23.47 & 27.00 & 25.59 & 30.50 & 29.63 & 0.784 & 0.692 & 0.756 & 0.776 & 0.841 & 0.853 & 0.236 & 0.217 & 0.253 & 0.220 & 0.247 & \cg 0.244\\
\textsc{Split} + \textit{ExtADC}$^*$ & 27.53 & 23.33 & 27.41 & 26.64 & 30.53 & 29.72 & 0.793 & 0.700 & \cg 0.765 & 0.796 & \cg 0.846 & \cg 0.857 & \cg 0.228 & 0.208 & \cg 0.238 & \cg 0.202 & \cg 0.245 & 0.245\\
\textsc{Joint} + \textit{SIG} & \cred 26.37 & \cred 22.90 & \cred 26.38 & \cred 25.01 & 29.41 & 28.14 & \cred 0.727 & \cred 0.637 & \cred 0.702 & \cred 0.704 & \cred 0.798 & \cred 0.796 & \cred 0.325 & 0.293 & \cred 0.345 & \cred 0.326 & \cred 0.336 & \cred 0.327\\
\textsc{Joint} + \textit{SIG} + \textit{ExtADC} & 26.53 & 23.52 & 26.51 & 25.29 & \cred 29.26 & \cred 28.05 & 0.737 & 0.656 & 0.709 & 0.715 & 0.804 & 0.802 & 0.308 & 0.272 & 0.328 & 0.308 & 0.319 & 0.311\\
\textsc{Joint} + \textit{SIG} + \textit{MSAD} & 26.65 & 23.28 & 26.56 & 25.36 & 29.62 & 28.45 & 0.742 & 0.661 & 0.713 & 0.724 & 0.807 & 0.804 & 0.270 & 0.240 & 0.287 & 0.268 & 0.281 & 0.273\\
\textsc{Joint} + \textit{SIG} + \textit{SpecDelay} & 26.77 & 23.71 & 26.49 & 25.46 & 29.62 & 28.55 & 0.756 & 0.702 & 0.717 & 0.734 & 0.816 & 0.812 & 0.260 & 0.212 & 0.282 & 0.264 & 0.274 & 0.269\\
\textsc{Joint} + \textit{SIG} + \textit{SpecDelay} + \textit{ExtADC} & 26.91 & 23.86 & 26.63 & 25.65 & 29.80 & 28.62 & 0.757 & 0.702 & 0.718 & 0.737 & 0.817 & 0.813 & 0.259 & 0.213 & 0.280 & 0.261 & 0.272 & 0.268\\
\textsc{Joint} + \textit{SIG} + \textit{SpecDelay} + \textit{MSAD} & 26.99 & \cg 24.22 & 26.63 & 25.74 & 29.68 & 28.67 & 0.767 & \cg 0.719 & 0.725 & 0.749 & 0.822 & 0.820 & 0.242 & \cg 0.198 & 0.262 & 0.242 & 0.257 & 0.252\\
\textsc{Joint} & 27.12 & 23.13 & 27.17 & 26.07 & 30.04 & 29.17 & 0.757 & 0.639 & 0.739 & 0.754 & 0.822 & 0.831 & 0.320 & \cred 0.312 & 0.331 & 0.305 & 0.330 & 0.321\\
\textsc{Joint} + \textit{ExtADC} & 27.12 & 23.23 & 27.17 & 26.11 & 30.00 & 29.08 & 0.758 & 0.642 & 0.741 & 0.756 & 0.822 & 0.831 & 0.317 & 0.306 & 0.329 & 0.301 & 0.328 & 0.319\\
\textsc{Joint} + \textit{MSAD} & \cg 27.77 & 23.85 & \cg 27.56 & 26.80 & \cg 30.62 & \cg 30.02 & 0.790 & 0.696 & 0.763 & 0.792 & 0.842 & 0.854 & 0.250 & 0.237 & 0.261 & 0.234 & 0.265 & 0.255\\
\textsc{Joint} + \textit{SpecDelay} & 27.53 & 23.80 & 27.30 & 26.41 & 30.36 & 29.76 & 0.782 & 0.699 & 0.753 & 0.778 & 0.837 & 0.844 & 0.256 & 0.225 & 0.272 & 0.249 & 0.273 & 0.262\\
\textsc{Joint} + \textit{SpecDelay} + \textit{ExtADC} & 27.62 & 23.98 & 27.40 & 26.65 & 30.41 & 29.65 & 0.786 & 0.702 & 0.757 & 0.783 & 0.839 & 0.848 & 0.257 & 0.228 & 0.272 & 0.248 & 0.273 & 0.263\\
\textsc{Joint} + \textit{SpecDelay} + \textit{MSAD} & \cg 27.77 & 24.14 & 27.53 & \cg 26.95 & 30.50 & 29.74 & \cg 0.796 & 0.718 & \cg 0.765 & \cg 0.797 & 0.845 & 0.855 & 0.239 & 0.213 & 0.253 & 0.226 & 0.256 & 0.247\\
\bottomrule
\end{tabular}
}

\end{table*}

\clearpage

\begin{figure*}[h!]
    \centering
        \includegraphics[width=\textwidth]{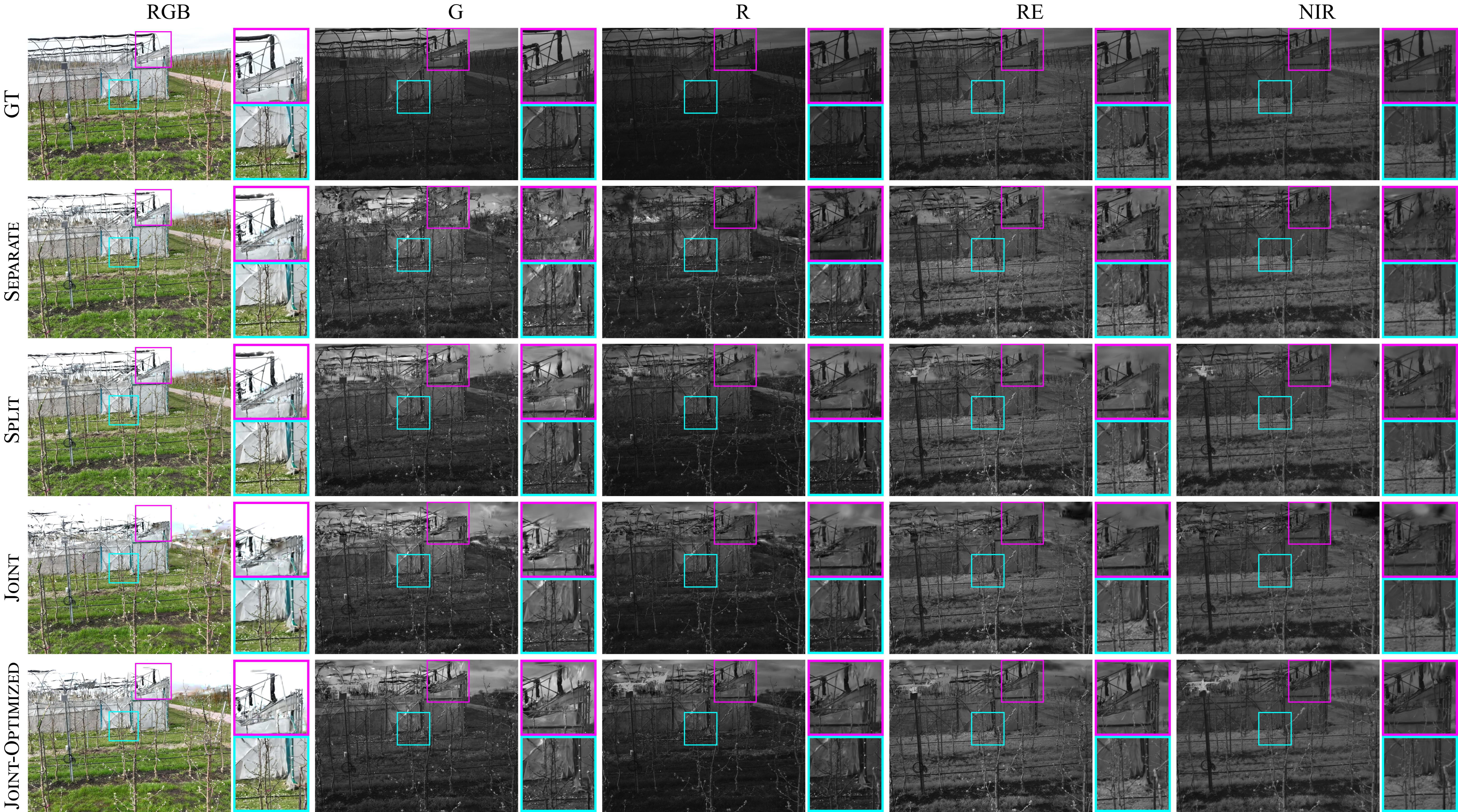}

    \caption{Visual comparison on the \textsc{Bud Swelling} scene with all available spectral bands.}
    \label{fig:comp_images_bud_swelling}
\end{figure*}

\begin{table*}[h!]

\caption[]{Comparison on \textsc{Bud Swelling} scene. The colors indicate the \colorbox{green!40}{best} and \colorbox{red!40}{worst} results.}

\tiny
\centering

\resizebox{\textwidth}{!}{
\begin{tabular}{ l | c c c c c c | c c c c c c | c c c c c c }
\toprule
& \multicolumn{6}{c|}{PSNR $(\uparrow)$} & \multicolumn{6}{c|}{SSIM $(\uparrow)$} & \multicolumn{6}{c}{LPIPS $(\downarrow)$}\\
\midrule
& All & RGB & G & R & RE & NIR & All & RGB & G & R & RE & NIR & All & RGB & G & R & RE & NIR\\
\midrule
\textsc{Separate} & 22.95 & 19.59 & \cred 21.10 & \cred 24.60 & 23.10 & 26.35 & 0.709 & 0.658 & 0.601 & \cred 0.743 & 0.727 & 0.818 & 0.263 & 0.182 & 0.322 & \cred 0.334 & 0.235 & \cred 0.241\\
\textsc{Split} & 23.99 & 19.59 & 22.03 & \cg 26.74 & 24.34 & 27.26 & 0.758 & 0.658 & 0.668 & 0.837 & 0.778 & 0.849 & 0.191 & 0.182 & 0.246 & \cg 0.190 & \cg 0.175 & \cg 0.163\\
\textsc{Split} + \textit{ExtADC}$^*$ & \cg 24.42 & 20.15 & 22.97 & \cg 26.74 & \cg 24.55 & \cg 27.71 & 0.765 & 0.676 & 0.685 & 0.837 & 0.778 & \cg 0.851 & \cg 0.190 & 0.166 & \cg 0.238 & 0.196 & 0.177 & 0.171\\
\textsc{Joint} + \textit{SIG} & \cred 22.50 & 18.41 & 21.73 & 25.00 & \cred 22.26 & \cred 25.09 & 0.665 & 0.555 & 0.585 & 0.755 & 0.671 & 0.760 & 0.274 & 0.256 & \cred 0.331 & 0.290 & \cred 0.255 & 0.238\\
\textsc{Joint} + \textit{SIG} + \textit{ExtADC} & 22.65 & 18.70 & 21.85 & 25.17 & 22.38 & 25.17 & 0.676 & 0.574 & 0.594 & 0.764 & 0.683 & 0.766 & 0.257 & 0.238 & 0.313 & 0.271 & 0.239 & 0.226\\
\textsc{Joint} + \textit{SIG} + \textit{MSAD} & 22.58 & 18.23 & 21.95 & 25.36 & \cred 22.26 & 25.10 & \cred 0.658 & \cred 0.544 & \cred 0.583 & 0.750 & \cred 0.665 & \cred 0.749 & 0.252 & 0.241 & 0.296 & 0.262 & 0.235 & 0.225\\
\textsc{Joint} + \textit{SIG} + \textit{SpecDelay} & 23.24 & 20.07 & 22.09 & 25.61 & 22.92 & 25.51 & 0.722 & 0.670 & 0.622 & 0.795 & 0.729 & 0.796 & 0.206 & 0.170 & 0.265 & 0.219 & 0.192 & 0.187\\
\textsc{Joint} + \textit{SIG} + \textit{SpecDelay} + \textit{ExtADC} & 23.26 & 20.08 & 22.12 & 25.67 & 22.92 & 25.51 & 0.722 & 0.669 & 0.621 & 0.795 & 0.728 & 0.795 & 0.209 & 0.174 & 0.267 & 0.222 & 0.194 & 0.189\\
\textsc{Joint} + \textit{SIG} + \textit{SpecDelay} + \textit{MSAD} & 23.43 & \cg 20.33 & 22.25 & 25.66 & 23.17 & 25.75 & 0.732 & 0.683 & 0.631 & 0.801 & 0.740 & 0.806 & 0.198 & \cg 0.165 & 0.254 & 0.212 & 0.182 & 0.176\\
\textsc{Joint} & 23.00 & \cred 17.96 & 22.15 & 25.88 & 23.00 & 26.03 & 0.703 & 0.562 & 0.639 & 0.791 & 0.719 & 0.805 & \cred 0.276 & \cred 0.284 & 0.328 & 0.279 & 0.254 & 0.235\\
\textsc{Joint} + \textit{ExtADC} & 23.20 & 18.45 & 22.13 & 25.76 & 23.31 & 26.33 & 0.712 & 0.580 & 0.647 & 0.798 & 0.728 & 0.809 & 0.262 & 0.267 & 0.315 & 0.264 & 0.240 & 0.224\\
\textsc{Joint} + \textit{MSAD} & 23.88 & 19.11 & 22.72 & 26.39 & 24.01 & 27.16 & 0.735 & 0.621 & 0.667 & 0.813 & 0.751 & 0.826 & 0.225 & 0.225 & 0.275 & 0.228 & 0.206 & 0.192\\
\textsc{Joint} + \textit{SpecDelay} & 24.20 & 20.10 & 23.00 & 26.56 & 24.25 & 27.06 & 0.760 & 0.670 & 0.683 & 0.833 & 0.774 & 0.842 & 0.202 & 0.186 & 0.258 & 0.207 & 0.186 & 0.173\\
\textsc{Joint} + \textit{SpecDelay} + \textit{ExtADC} & 24.12 & 20.12 & 22.99 & 26.44 & 24.16 & 26.91 & 0.760 & 0.669 & 0.684 & 0.832 & 0.772 & 0.843 & 0.209 & 0.197 & 0.265 & 0.213 & 0.193 & 0.179\\
\textsc{Joint} + \textit{SpecDelay} + \textit{MSAD} & 24.29 & 20.26 & \cg 23.01 & 26.62 & 24.35 & 27.19 & \cg 0.769 & \cg 0.686 & \cg 0.692 & \cg 0.838 & \cg 0.782 & 0.850 & 0.195 & 0.180 & 0.248 & 0.199 & 0.179 & 0.167\\
\bottomrule
\end{tabular}
}

\end{table*}

\clearpage

\begin{figure*}[h!]
    \centering
        \includegraphics[width=\textwidth]{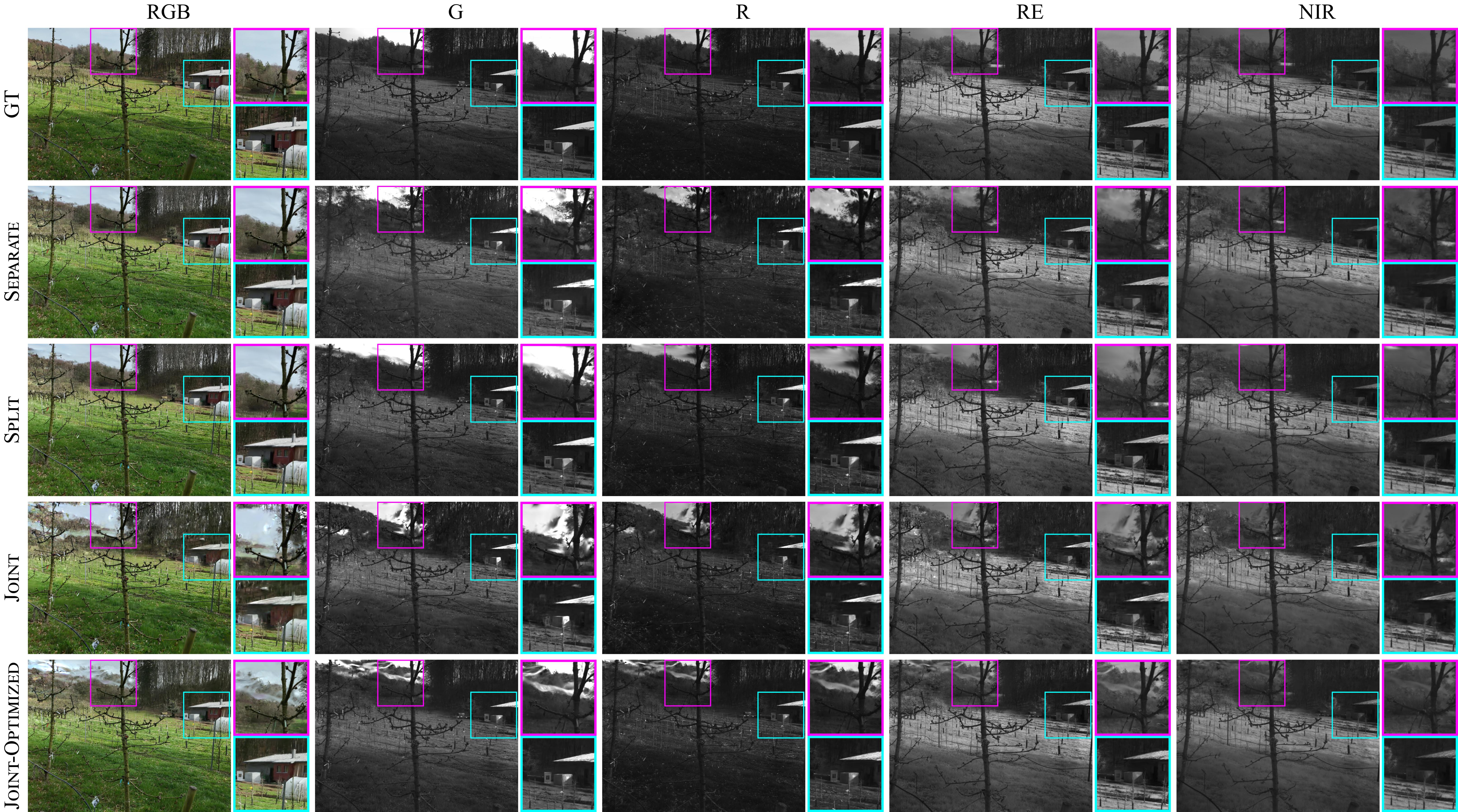}

    \caption{Visual comparison on the \textsc{Single Tree} scene with all available spectral bands.}
    \label{fig:comp_images_single_tree_2}
\end{figure*}

\begin{table*}[h!]

\caption[]{Comparison on \textsc{Single Tree} scene. The colors indicate the \colorbox{green!40}{best} and \colorbox{red!40}{worst} results.}

\tiny
\centering

\resizebox{\textwidth}{!}{
\begin{tabular}{ l | c c c c c c | c c c c c c | c c c c c c }
\toprule
& \multicolumn{6}{c|}{PSNR $(\uparrow)$} & \multicolumn{6}{c|}{SSIM $(\uparrow)$} & \multicolumn{6}{c}{LPIPS $(\downarrow)$}\\
\midrule
& All & RGB & G & R & RE & NIR & All & RGB & G & R & RE & NIR & All & RGB & G & R & RE & NIR\\
\midrule
\textsc{Separate} & 22.53 & 19.38 & 22.07 & 23.44 & 22.36 & 25.41 & 0.677 & 0.490 & 0.598 & 0.731 & 0.712 & 0.853 & 0.359 & 0.263 & 0.378 & 0.372 & 0.357 & \cred 0.427\\
\textsc{Split} & 24.28 & 19.38 & 23.92 & 25.53 & 24.88 & 27.67 & 0.706 & 0.490 & 0.639 & 0.782 & 0.756 & 0.861 & 0.257 & 0.263 & 0.284 & 0.245 & 0.237 & 0.255\\
\textsc{Split} + \textit{ExtADC}$^*$ & 24.21 & 19.58 & 23.78 & 25.29 & 24.93 & 27.46 & 0.708 & 0.502 & 0.638 & 0.779 & 0.756 & 0.864 & 0.266 & 0.252 & 0.295 & 0.259 & 0.251 & 0.275\\
\textsc{Joint} + \textit{SIG} & \cred 21.17 & 16.59 & \cred 20.77 & \cred 21.80 & \cred 21.91 & 24.78 & 0.588 & \cred 0.281 & 0.564 & 0.667 & 0.636 & 0.795 & \cred 0.464 & 0.505 & \cred 0.500 & \cred 0.465 & 0.437 & 0.415\\
\textsc{Joint} + \textit{SIG} + \textit{ExtADC} & 21.85 & 17.10 & 21.55 & 22.74 & 22.47 & 25.41 & 0.610 & 0.333 & 0.573 & 0.685 & 0.656 & 0.801 & 0.393 & 0.419 & 0.428 & 0.394 & 0.368 & 0.356\\
\textsc{Joint} + \textit{SIG} + \textit{MSAD} & 21.48 & 16.80 & 21.32 & 22.22 & 22.06 & 24.98 & \cred 0.585 & 0.321 & \cred 0.545 & \cred 0.658 & \cred 0.625 & \cred 0.778 & 0.362 & 0.378 & 0.387 & 0.369 & 0.340 & 0.338\\
\textsc{Joint} + \textit{SIG} + \textit{SpecDelay} & 24.57 & 19.62 & 24.15 & 25.66 & 24.92 & 28.49 & 0.694 & 0.499 & 0.627 & 0.759 & 0.735 & 0.853 & 0.255 & 0.255 & 0.294 & 0.262 & 0.235 & 0.231\\
\textsc{Joint} + \textit{SIG} + \textit{SpecDelay} + \textit{ExtADC} & 24.34 & 19.41 & 23.97 & 25.29 & 24.67 & 28.35 & 0.694 & 0.495 & 0.627 & 0.758 & 0.735 & 0.853 & 0.260 & 0.262 & 0.297 & 0.268 & 0.239 & 0.236\\
\textsc{Joint} + \textit{SIG} + \textit{SpecDelay} + \textit{MSAD} & 25.05 & 20.27 & 24.34 & 26.03 & 25.44 & 29.16 & 0.711 & 0.530 & 0.638 & 0.775 & 0.751 & 0.861 & \cg 0.241 & \cg 0.242 & \cg 0.278 & \cg 0.244 & \cg 0.223 & \cg 0.220\\
\textsc{Joint} & 21.35 & \cred 16.47 & 21.26 & 22.56 & 21.95 & \cred 24.50 & 0.634 & 0.344 & 0.601 & 0.714 & 0.687 & 0.825 & 0.462 & \cred 0.510 & \cred 0.500 & 0.438 & \cred 0.438 & 0.423\\
\textsc{Joint} + \textit{ExtADC} & 22.21 & 17.24 & 21.87 & 23.29 & 22.64 & 26.01 & 0.648 & 0.378 & 0.605 & 0.727 & 0.696 & 0.833 & 0.412 & 0.450 & 0.452 & 0.389 & 0.391 & 0.375\\
\textsc{Joint} + \textit{MSAD} & 23.75 & 18.73 & 23.20 & 24.73 & 24.35 & 27.74 & 0.686 & 0.450 & 0.632 & 0.759 & 0.736 & 0.855 & 0.325 & 0.361 & 0.362 & 0.306 & 0.301 & 0.293\\
\textsc{Joint} + \textit{SpecDelay} & 24.62 & 19.69 & 24.30 & 25.60 & 25.13 & 28.41 & 0.714 & 0.513 & 0.650 & 0.781 & 0.760 & 0.869 & 0.265 & 0.278 & 0.306 & 0.258 & 0.244 & 0.239\\
\textsc{Joint} + \textit{SpecDelay} + \textit{ExtADC} & 24.17 & 19.56 & 23.62 & 24.94 & 24.90 & 27.81 & 0.715 & 0.512 & 0.652 & 0.780 & 0.763 & 0.868 & 0.285 & 0.298 & 0.330 & 0.276 & 0.262 & 0.257\\
\textsc{Joint} + \textit{SpecDelay} + \textit{MSAD} & \cg 25.27 & \cg 20.39 & \cg 24.48 & \cg 26.26 & \cg 25.86 & \cg 29.33 & \cg 0.730 & \cg 0.538 & \cg 0.662 & \cg 0.798 & \cg 0.775 & \cg 0.875 & 0.260 & 0.273 & 0.303 & 0.247 & 0.241 & 0.237\\
\bottomrule
\end{tabular}
}

\end{table*}

\clearpage

\begin{figure*}[h!]
    \centering
        \includegraphics[width=\textwidth]{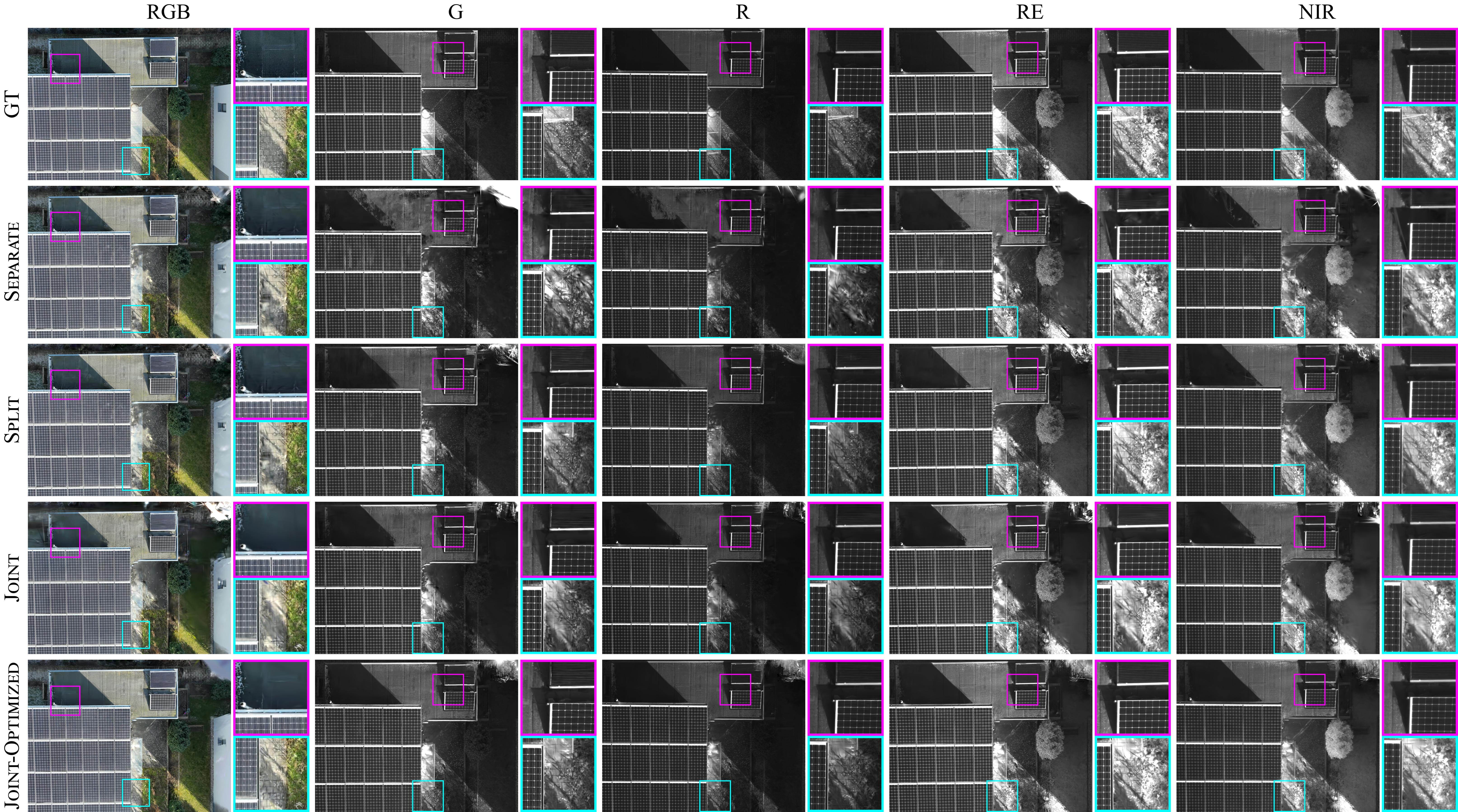}

    \caption{Visual comparison on the \textsc{Solar} scene with all available spectral bands.}
    \label{fig:comp_images_solar}
\end{figure*}

\begin{table*}[h!]

\caption[]{Comparison on \textsc{Solar} scene. The colors indicate the \colorbox{green!40}{best} and \colorbox{red!40}{worst} results.}

\tiny
\centering

\resizebox{\textwidth}{!}{
\begin{tabular}{ l | c c c c c c | c c c c c c | c c c c c c }
\toprule
& \multicolumn{6}{c|}{PSNR $(\uparrow)$} & \multicolumn{6}{c|}{SSIM $(\uparrow)$} & \multicolumn{6}{c}{LPIPS $(\downarrow)$}\\
\midrule
& All & RGB & G & R & RE & NIR & All & RGB & G & R & RE & NIR & All & RGB & G & R & RE & NIR\\
\midrule
\textsc{Separate} & 23.24 & 21.17 & 22.45 & 25.99 & 22.17 & 24.42 & 0.809 & 0.735 & 0.793 & 0.857 & 0.808 & 0.852 & 0.231 & 0.234 & 0.248 & 0.204 & 0.237 & \cred 0.234\\
\textsc{Split} & 23.76 & 21.17 & 22.99 & 26.44 & 22.94 & 25.28 & 0.817 & 0.735 & 0.804 & 0.866 & 0.820 & 0.860 & 0.198 & 0.234 & 0.211 & 0.159 & 0.199 & 0.188\\
\textsc{Split} + \textit{ExtADC}$^*$ & \cg 23.96 & \cg 21.79 & \cg 23.41 & 26.52 & 22.79 & 25.28 & 0.822 & 0.749 & \cg 0.811 & \cg 0.869 & 0.821 & 0.861 & \cg 0.193 & \cg 0.211 & \cg 0.205 & \cg 0.157 & 0.201 & 0.191\\
\textsc{Joint} + \textit{SIG} & 22.54 & 20.70 & \cred 21.31 & 25.09 & 21.76 & 23.85 & \cred 0.772 & 0.699 & \cred 0.731 & \cred 0.831 & \cred 0.777 & \cred 0.822 & \cred 0.253 & 0.287 & \cred 0.282 & \cred 0.211 & \cred 0.252 & 0.232\\
\textsc{Joint} + \textit{SIG} + \textit{ExtADC} & \cred 22.36 & 20.67 & 21.32 & \cred 24.87 & \cred 21.50 & \cred 23.45 & 0.775 & 0.703 & 0.738 & 0.833 & 0.778 & 0.823 & 0.246 & 0.277 & 0.272 & 0.206 & 0.247 & 0.229\\
\textsc{Joint} + \textit{SIG} + \textit{MSAD} & 22.78 & 20.91 & 21.67 & 25.33 & 21.96 & 24.03 & 0.782 & 0.709 & 0.746 & 0.838 & 0.785 & 0.829 & 0.228 & 0.255 & 0.252 & 0.190 & 0.229 & 0.214\\
\textsc{Joint} + \textit{SIG} + \textit{SpecDelay} & 22.89 & 21.57 & 21.57 & 25.19 & 22.06 & 24.07 & 0.789 & 0.746 & 0.743 & 0.836 & 0.791 & 0.829 & 0.217 & 0.221 & 0.248 & 0.188 & 0.219 & 0.210\\
\textsc{Joint} + \textit{SIG} + \textit{SpecDelay} + \textit{ExtADC} & 22.92 & 21.56 & 21.61 & 25.22 & 22.07 & 24.13 & 0.792 & 0.745 & 0.749 & 0.840 & 0.794 & 0.834 & 0.216 & 0.224 & 0.245 & 0.185 & 0.217 & 0.207\\
\textsc{Joint} + \textit{SIG} + \textit{SpecDelay} + \textit{MSAD} & 23.09 & 21.67 & 21.82 & 25.41 & 22.27 & 24.28 & 0.798 & 0.749 & 0.756 & 0.843 & 0.800 & 0.838 & 0.208 & 0.217 & 0.237 & 0.178 & 0.209 & 0.198\\
\textsc{Joint} & 23.67 & \cred 20.63 & 23.18 & 26.40 & 22.92 & 25.24 & 0.801 & \cred 0.695 & 0.790 & 0.856 & 0.810 & 0.854 & 0.244 & \cred 0.306 & 0.260 & 0.196 & 0.244 & 0.216\\
\textsc{Joint} + \textit{ExtADC} & 23.49 & 20.68 & 22.93 & 26.23 & 22.66 & 24.97 & 0.806 & 0.707 & 0.793 & 0.860 & 0.813 & 0.856 & 0.234 & 0.294 & 0.250 & 0.186 & 0.234 & 0.207\\
\textsc{Joint} + \textit{MSAD} & 23.64 & 21.05 & 23.03 & 26.21 & 22.80 & 25.09 & 0.813 & 0.728 & 0.799 & 0.862 & 0.819 & 0.859 & 0.207 & 0.251 & 0.222 & 0.166 & 0.208 & 0.189\\
\textsc{Joint} + \textit{SpecDelay} & 23.77 & 21.34 & 22.86 & 26.33 & 23.05 & 25.29 & 0.817 & 0.746 & 0.796 & 0.862 & 0.823 & 0.860 & 0.201 & 0.230 & 0.224 & 0.166 & 0.203 & 0.186\\
\textsc{Joint} + \textit{SpecDelay} + \textit{ExtADC} & 23.92 & 21.53 & 23.07 & \cg 26.56 & 23.08 & 25.36 & 0.820 & 0.746 & 0.800 & 0.865 & 0.826 & 0.862 & 0.203 & 0.235 & 0.224 & 0.166 & 0.205 & 0.187\\
\textsc{Joint} + \textit{SpecDelay} + \textit{MSAD} & 23.95 & 21.66 & 23.13 & 26.36 & \cg 23.18 & \cg 25.43 & \cg 0.824 & \cg 0.752 & 0.805 & 0.867 & \cg 0.830 & \cg 0.865 & 0.194 & 0.225 & 0.213 & 0.158 & \cg 0.194 & \cg 0.178\\
\bottomrule
\end{tabular}
}

\end{table*}

\clearpage

\begin{figure*}[h!]
    \centering
    \includegraphics[width=\textwidth]{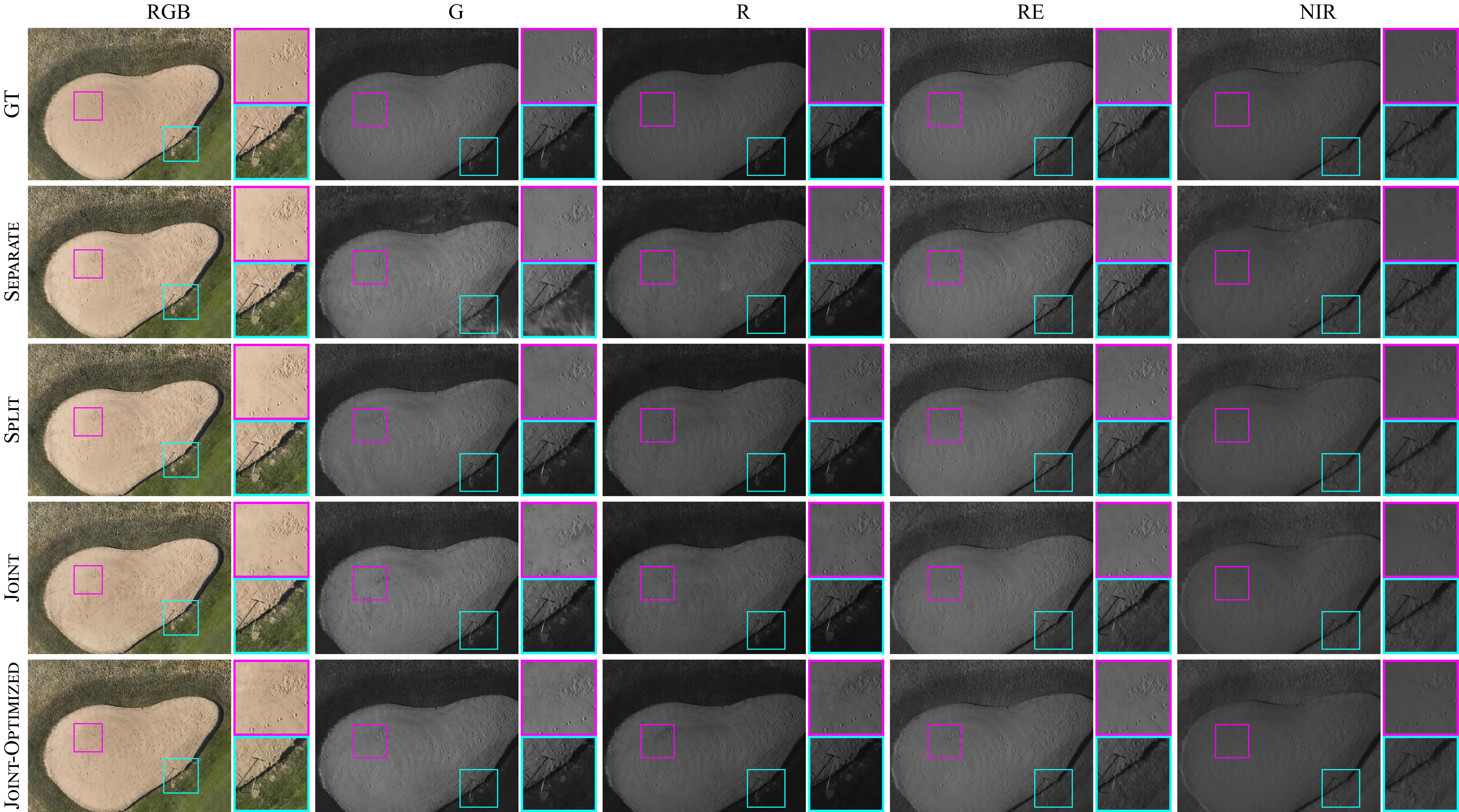}

    \caption{Visual comparison on the \textsc{Golf} scene with all available spectral bands.}
    \label{fig:comp_images_golf}
\end{figure*}

\begin{table*}[h!]

\caption[]{Comparison on \textsc{Golf} scene. The colors indicate the \colorbox{green!40}{best} and \colorbox{red!40}{worst} results.}

\tiny
\centering

\resizebox{\textwidth}{!}{
\begin{tabular}{ l | c c c c c c | c c c c c c | c c c c c c }
\toprule
& \multicolumn{6}{c|}{PSNR $(\uparrow)$} & \multicolumn{6}{c|}{SSIM $(\uparrow)$} & \multicolumn{6}{c}{LPIPS $(\downarrow)$}\\
\midrule
& All & RGB & G & R & RE & NIR & All & RGB & G & R & RE & NIR & All & RGB & G & R & RE & NIR\\
\midrule
\textsc{Separate} & \cred 27.62 & 24.98 & \cred 26.04 & \cred 28.89 & \cred 26.43 & \cred 31.77 & 0.812 & 0.710 & 0.782 & 0.856 & 0.807 & 0.904 & 0.431 & 0.303 & \cred 0.497 & \cred 0.478 & 0.455 & \cred 0.424\\
\textsc{Split} & 31.30 & 24.98 & 31.17 & 32.62 & \cg 31.66 & 36.09 & \cg 0.854 & 0.710 & 0.853 & 0.902 & \cg 0.863 & \cg 0.945 & \cg 0.233 & 0.303 & 0.248 & 0.225 & \cg 0.215 & \cg 0.174\\
\textsc{Split} + \textit{ExtADC}$^*$ & 30.02 & 24.77 & 30.01 & 31.66 & 27.55 & 36.09 & 0.849 & 0.712 & 0.846 & 0.900 & 0.844 & 0.944 & 0.258 & \cg 0.294 & 0.275 & 0.265 & 0.259 & 0.199\\
\textsc{Joint} + \textit{SIG} & 28.02 & 22.58 & 28.01 & 29.13 & 28.07 & 32.78 & 0.745 & 0.482 & 0.775 & 0.842 & 0.757 & 0.881 & \cred 0.465 & \cred 0.572 & 0.459 & 0.403 & \cred 0.482 & 0.405\\
\textsc{Joint} + \textit{SIG} + \textit{ExtADC} & 27.98 & \cred 22.35 & 28.02 & 29.44 & 28.04 & 32.54 & \cred 0.736 & \cred 0.465 & \cred 0.766 & \cred 0.840 & \cred 0.748 & \cred 0.874 & 0.449 & 0.544 & 0.445 & 0.387 & 0.469 & 0.395\\
\textsc{Joint} + \textit{SIG} + \textit{MSAD} & 28.91 & 24.18 & 28.56 & 29.95 & 28.58 & 33.78 & 0.789 & 0.614 & 0.797 & 0.858 & 0.787 & 0.901 & 0.327 & 0.407 & 0.330 & 0.288 & 0.332 & 0.273\\
\textsc{Joint} + \textit{SIG} + \textit{SpecDelay} & 29.17 & 24.68 & 28.69 & 30.07 & 28.89 & 34.03 & 0.831 & 0.712 & 0.826 & 0.882 & 0.823 & 0.920 & 0.243 & 0.299 & 0.254 & 0.226 & 0.237 & 0.192\\
\textsc{Joint} + \textit{SIG} + \textit{SpecDelay} + \textit{ExtADC} & 29.23 & 24.99 & 28.73 & 29.95 & 28.86 & 34.09 & 0.831 & 0.713 & 0.826 & 0.881 & 0.821 & 0.920 & 0.244 & 0.299 & 0.255 & 0.228 & 0.239 & 0.194\\
\textsc{Joint} + \textit{SIG} + \textit{SpecDelay} + \textit{MSAD} & 29.25 & 24.94 & 28.73 & 29.95 & 28.96 & 34.18 & 0.833 & 0.716 & 0.828 & 0.882 & 0.825 & 0.922 & 0.240 & 0.296 & 0.250 & 0.225 & 0.233 & 0.189\\
\textsc{Joint} & 29.27 & 23.48 & 29.37 & 31.00 & 29.06 & 33.87 & 0.813 & 0.602 & 0.832 & 0.890 & 0.828 & 0.924 & 0.395 & 0.518 & 0.393 & 0.331 & 0.408 & 0.316\\
\textsc{Joint} + \textit{ExtADC} & 30.49 & 24.26 & 30.59 & 32.16 & 30.56 & 35.39 & 0.820 & 0.632 & 0.833 & 0.889 & 0.830 & 0.927 & 0.361 & 0.467 & 0.365 & 0.308 & 0.368 & 0.291\\
\textsc{Joint} + \textit{MSAD} & 28.68 & 23.44 & 29.00 & 30.46 & 28.24 & 32.66 & 0.815 & 0.624 & 0.835 & 0.880 & 0.826 & 0.923 & 0.375 & 0.477 & 0.378 & 0.317 & 0.387 & 0.311\\
\textsc{Joint} + \textit{SpecDelay} & 30.65 & 24.63 & 31.03 & 32.34 & 30.83 & 34.85 & 0.850 & 0.713 & 0.852 & 0.899 & 0.857 & 0.939 & 0.250 & 0.328 & 0.264 & 0.229 & 0.232 & 0.188\\
\textsc{Joint} + \textit{SpecDelay} + \textit{ExtADC} & 30.87 & 24.96 & 30.77 & 32.04 & 31.12 & 35.98 & 0.853 & 0.713 & \cg 0.854 & 0.903 & 0.859 & 0.942 & 0.245 & 0.327 & 0.258 & 0.223 & 0.227 & 0.182\\
\textsc{Joint} + \textit{SpecDelay} + \textit{MSAD} & \cg 31.58 & \cg 25.06 & \cg 31.75 & \cg 33.48 & 31.65 & \cg 36.46 & \cg 0.854 & \cg 0.719 & \cg 0.854 & \cg 0.905 & 0.859 & 0.942 & 0.235 & 0.313 & \cg 0.247 & \cg 0.218 & \cg 0.215 & 0.175\\
\bottomrule
\end{tabular}
}

\end{table*}

\clearpage

\begin{table*}[h!]
\caption{Number of Gaussian splats required for the \textsc{Garden} scene.}
\centering

\begin{tabular}{ l | c c c c c }
\toprule
 & RGB & G & R & RE & NIR \\
\midrule
\textsc{Separate}                                 & 1,149,797 & 1,079,264 & 1,209,323 &   750,358 &   714,917 \\
\textsc{Split}                                    & 1,149,797 & 1,195,089 & 1,183,735 & 1,200,807 & 1,183,349 \\
\textsc{Split} + \textit{ExtADC}$^*$              & 1,278,153 & 1,421,841 & 1,566,597 & 1,129,455 & 1,096,379 \\
\textsc{Joint} + \textit{SIG}                     & 1,029,761 & 1,029,761 & 1,029,761 & 1,029,761 & 1,029,761 \\
\textsc{Joint} + \textit{SIG} + \textit{ExtADC}   & 1,160,343 & 1,160,343 & 1,160,343 & 1,160,343 & 1,160,343 \\
\textsc{Joint} + \textit{SIG} + \textit{MSAD}     & 2,164,098 & 2,164,098 & 2,164,098 & 2,164,098 & 2,164,098 \\
\textsc{Joint} + \textit{SIG} + \textit{SpecDelay}& 1,189,156 & 1,189,156 & 1,189,156 & 1,189,156 & 1,189,156 \\
\textsc{Joint} + \textit{SIG} + \textit{SpecDelay} + \textit{ExtADC} & 1,376,332 & 1,376,332 & 1,376,332 & 1,376,332 & 1,376,332 \\
\textsc{Joint} + \textit{SIG} + \textit{SpecDelay} + \textit{MSAD}   & 1,987,389 & 1,987,389 & 1,987,389 & 1,987,389 & 1,987,389 \\
\textsc{Joint}                                    &   948,810 &   948,810 &   948,810 &   948,810 &   948,810 \\
\textsc{Joint} + \textit{ExtADC}                  & 1,099,171 & 1,099,171 & 1,099,171 & 1,099,171 & 1,099,171 \\
\textsc{Joint} + \textit{MSAD}                    & 1,920,698 & 1,920,698 & 1,920,698 & 1,920,698 & 1,920,698 \\
\textsc{Joint} + \textit{SpecDelay}               & 1,191,627 & 1,191,627 & 1,191,627 & 1,191,627 & 1,191,627 \\
\textsc{Joint} + \textit{SpecDelay} + \textit{ExtADC} & 1,285,691 & 1,285,691 & 1,285,691 & 1,285,691 & 1,285,691 \\
\textsc{Joint} + \textit{SpecDelay} + \textit{MSAD}   & 1,916,049 & 1,916,049 & 1,916,049 & 1,916,049 & 1,916,049 \\
\bottomrule
\end{tabular}
\end{table*}

\begin{table*}[h!]
\caption{Number of Gaussian splats required for the \textsc{Lake} scene.}
\centering

\begin{tabular}{ l | c c c c c }
\toprule
 & RGB & G & R & RE & NIR \\
\midrule
\textsc{Separate}                                 & 2,613,207 & 2,609,129 & 1,445,470 & 2,637,019 & 1,917,627 \\
\textsc{Split}                                    & 2,613,207 & 2,874,413 & 2,883,637 & 2,934,146 & 3,602,592 \\
\textsc{Split} + \textit{ExtADC}$^*$              & 3,395,208 & 3,250,713 & 2,490,202 & 5,645,837 & 2,852,237 \\
\textsc{Joint} + \textit{SIG}                     & 1,448,340 & 1,448,340 & 1,448,340 & 1,448,340 & 1,448,340 \\
\textsc{Joint} + \textit{SIG} + \textit{ExtADC}   & 2,119,385 & 2,119,385 & 2,119,385 & 2,119,385 & 2,119,385 \\
\textsc{Joint} + \textit{SIG} + \textit{MSAD}     & 4,726,836 & 4,726,836 & 4,726,836 & 4,726,836 & 4,726,836 \\
\textsc{Joint} + \textit{SIG} + \textit{SpecDelay}& 3,207,813 & 3,207,813 & 3,207,813 & 3,207,813 & 3,207,813 \\
\textsc{Joint} + \textit{SIG} + \textit{SpecDelay} + \textit{ExtADC} & 3,838,578 & 3,838,578 & 3,838,578 & 3,838,578 & 3,838,578 \\
\textsc{Joint} + \textit{SIG} + \textit{SpecDelay} + \textit{MSAD}   & 5,522,547 & 5,522,547 & 5,522,547 & 5,522,547 & 5,522,547 \\
\textsc{Joint}                                    & 2,310,506 & 2,310,506 & 2,310,506 & 2,310,506 & 2,310,506 \\
\textsc{Joint} + \textit{ExtADC}                  & 3,376,816 & 3,376,816 & 3,376,816 & 3,376,816 & 3,376,816 \\
\textsc{Joint} + \textit{MSAD}                    & 4,859,472 & 4,859,472 & 4,859,472 & 4,859,472 & 4,859,472 \\
\textsc{Joint} + \textit{SpecDelay}               & 3,502,881 & 3,502,881 & 3,502,881 & 3,502,881 & 3,502,881 \\
\textsc{Joint} + \textit{SpecDelay} + \textit{ExtADC} & 3,326,236 & 3,326,236 & 3,326,236 & 3,326,236 & 3,326,236 \\
\textsc{Joint} + \textit{SpecDelay} + \textit{MSAD}   & 5,557,032 & 5,557,032 & 5,557,032 & 5,557,032 & 5,557,032 \\
\bottomrule
\end{tabular}

\end{table*}

\end{document}